\documentclass[letterpaper]{article}
\usepackage{tikz}
\usetikzlibrary{quantikz}
\usepackage{times}
\usepackage{epsfig}
\usepackage{graphicx}
\usepackage{amsmath}
\usepackage{amssymb}
\usepackage{dsfont}
\usepackage[toc,page]{appendix}
\usepackage{physics}
\usepackage[pagebackref,breaklinks,colorlinks]{hyperref}

\DeclareGraphicsExtensions{.pdf,.png}

\begin{document}

\title{Quantum Levenberg--Marquardt Algorithm for optimization in Bundle Adjustment}

\author{Luca Bernecker \and 
Andrea Idini
}
%
\date{
{\it
Division of Mathematical Physics, Department of Physics, LTH, \\ Lund University, P.O. Box 118, S-22100 Lund, Sweden}\\
andrea.idini@matfys.lth.se
}

\maketitle


\begin{abstract}
 \noindent 
In this paper we develop a quantum optimization algorithm and use it to solve the bundle adjustment problem with a simulated quantum computer. Bundle adjustment is the process of optimizing camera poses and sensor properties to best reconstruct the three-dimensional structure and viewing parameters. This problem is often solved using some implementation of the Levenberg--Marquardt algorithm. 
In this case we implement a quantum algorithm for solving the linear system of normal equations that calculates the optimization step in Levenberg--Marquardt. This procedure is the current bottleneck in the algorithmic complexity of bundle adjustment. The proposed quantum algorithm dramatically reduces the complexity of this operation with respect to the number of points.

We investigate 9 configurations of a toy-model for bundle adjustment, limited to 10 points and 2 cameras. This optimization problem is solved both by using the sparse Levenberg-Marquardt algorithm and our quantum implementation. The resulting solutions are presented, showing an improved rate of convergence, together with an analysis of the theoretical speed up and the probability of running the algorithm successfully on a current quantum computer. 

The presented quantum algorithm is a seminal implementation of using quantum computing algorithms in order to solve complex optimization problems in computer vision, in particular bundle adjustment, which offers several avenues of further investigations.
\end{abstract}

\section{Introduction}

Reconstructing the three--dimensional scene geometry from a set of sensor readings is a long standing problem in computer vision. Bundle adjustment (BA) is a fundamental problem in computer vision that pertains to 3D reconstruction. It is the process of jointly calibrating sensor parameters and the 3D geometric structure of the captured scene to produce the optimal reconstruction \cite{triggs1999}. Therefore, its solution has a great number of applications, e.g. large scale reconstruction, visual positioning systems \cite{liu2017creating}, robotics \cite{pradeep2014calibrating}, and even space exploration \cite{maimone2007two}.

The computational bottleneck of scene reconstruction from large amount of images consists in the optimization algorithm used in BA. Several optimization algorithms have been used to tackle this problem \cite{lourakis2005,chen2019bundle,peng2021scalable}. The Levenberg--Marquardt algorithm (LMA) is widely considered to be the most efficient for this problem. In its simplest implementation, LMA consists in a linear combination of gradient descent (linear) and Gauss-Newton (quadratic). For bundle adjustment, LMA with Schur complement trick has an algorithmic complexity scaling of the order of $O(m^3 + mn)$, with $m$ the number of cameras/features and $n$ the number of 3D points. The $O(m^3)$ scaling is due to the inversion of the Hessian matrix in order to find the solution to the linear problem and calculate the update step of the LMA \cite{hartley2003}.

In this paper, we develop a quantum computing algorithm to solve this bottleneck of the LMA, that is the linear problem of the Hessian inversion. At this stage, we employ the Harrow, Hassidim, and Lloyd (HHL) algorithm \cite{hhl2009}. HHL has, potentially, only a logarithmic scaling with respect to the size of the matrix \cite{dervovic2018quantum}, hence with respect to the number of features. In the following, we call this hybrid quantum computing implementation of the Levenberg--Marquardt algorithm QLMA.



In particular, we investigate this quantum algorithm application to bundle adjustment through the example of minimizing the re-projection error of 2 cameras and 10 points with 9 different test configurations. We make use of the HHL algorithm and develop the QLMA implementation. These results are then compared to the results of the sparse LMA. We conclude that the implementation has advantages not only in complexity scaling, but also in reaching an approximate solution in fewer steps. This result suggests that the linear approximations used for the quantum implementation might provide an advantage over the full solution for difficult problems. Furthermore, this study encourages the possibility of quantum computers as platform for the solution of bundle adjustment and other optimization problems in computer vision in the future.

\section{Related work}
\label{section2.1}

\textbf{Bundle Adjustment}. In a nutshell, bundle adjustment is a cost-minimization procedure. It is based on the minimization of the error in reconstruction by the consistency between different sensors information. Considering only image data, the key metric to minimize in BA is the re-projection error. BA is known to greatly impact the execution time of a large-scale reconstruction pipeline \cite{triggs1999,Das_2021_WACV}. Due to the possible applications of large scale deployment of this technique \cite{liu2017creating,pradeep2014calibrating}, several different approaches have been tried for efficient minimization \cite{lourakis2005,chen2019bundle}. Especially, recent interest involve multicore and GPU solutions \cite{agarwal2011,wu2011,eriksson2016,zhang2017,Demmel_2021_CVPR} and the development on specialized architectures such as graph processors \cite{Ortiz_2020_CVPR}.

The most often used algorithm for BA is Levenberg--Marquardt. That is a modified Gauss-Newton algorithm, based on a local linearization of the residuals obtained assuming small steps \cite{hartley2003,transtrum2012improvements}. 
Several open source implementations of bundle adjustment are available, e.g., SBA \cite{lourakis2009sba}, Theia \cite{theia-manual}, and of course Ceres \cite{ceres-solver}. The latter is widely considered the industry standard for bundle adjustment and non--linear optimization in general. All of these libraries are based on sparse implementation of LMA.

\textbf{Quantum Computing}. Quantum computing is the study of computational algorithms based on quantum bits (qubits or qbits). Differently from the familiar binary bits, qubits are vectors in a two-dimensional Hilbert space. Therefore, instead of just binary values the units of computation in quantum computing are vectors in a space over the complex field \cite{nielsen_chuang_2010}. Using this property, it was realized before any practical implementation that quantum computing is a new computational paradigm. It was found that quantum computers may hold a significant advantage in the scaling of a class of optimization problems noted as bounded error quantum polynomial time \cite{benioff1980computer,bernstein1997quantum}. This class of problems is assumed to be larger than the classical polynomial class, as notably popularized by John Preskill's famous term {\it quantum supremacy}  \cite{preskill2012quantum}. However, we want to stress that despite several results pointing to a dramatic advantage of specific implementations \cite{arute2019quantum}, quantum supremacy has not yet been formally demonstrated in terms of complexity classes.

It is generally difficult to implement quantum algorithms and demonstrate the scaling in practical settings on present, or even near--future, quantum devices if not for a limited set of algorithms and scenarios. This is due to noisy qubits, limited number and connectivity of qubits, and decoherence time. Therefore, hybrid quantum-classical optimization algorithms have been studied and implemented 
\cite{farhi2014quantum,peruzzo2014variational,wecker2015,McClean2016,ajegekar2020} to optimize different scientific problems in fields such as quantum chemistry and particle physics \cite{cao2019quantum, mott2017solving}. Recently, some quantum implementations of elements of a computer vision pipeline have been proposed and sometime tested, e.g. robust fitting \cite{Chin2020,doan2022hybrid}.
The objective of this paper is to extend the applications of quantum computing to the optimization problem of bundle adjustment.
In particular, we propose to solve the linear equation needed to find the linearized step of LMA using the quantum computing algorithm HHL with some specific approximations \cite{hhl2009}. 

Our QLMA implementation will be tested in a quantum simulator and compared to the classical sparse LMA. In 1997, LLoyd and Abrams demonstrated the possibility of an efficient simulation of quantum calculation \cite{Abrams1997}. This type of simulation is now applied in a module called Aer from IBM's quantum computing platform to enable prototyping quantum algorithms \cite{Qiskit-Textbook}.

\section{Methodology}
\label{method}

\subsection{Bundle-Adjustment}
Bundle  adjustment can be viewed to be a  large  sparse  geometric  parameter  estimation  problem.  In our case, the  parameters are the camera orientations and position that then determine the reconstruction and reprojections of the points.
 
\begin{figure}[h!]
\centering
\includegraphics[width = 7.8cm]{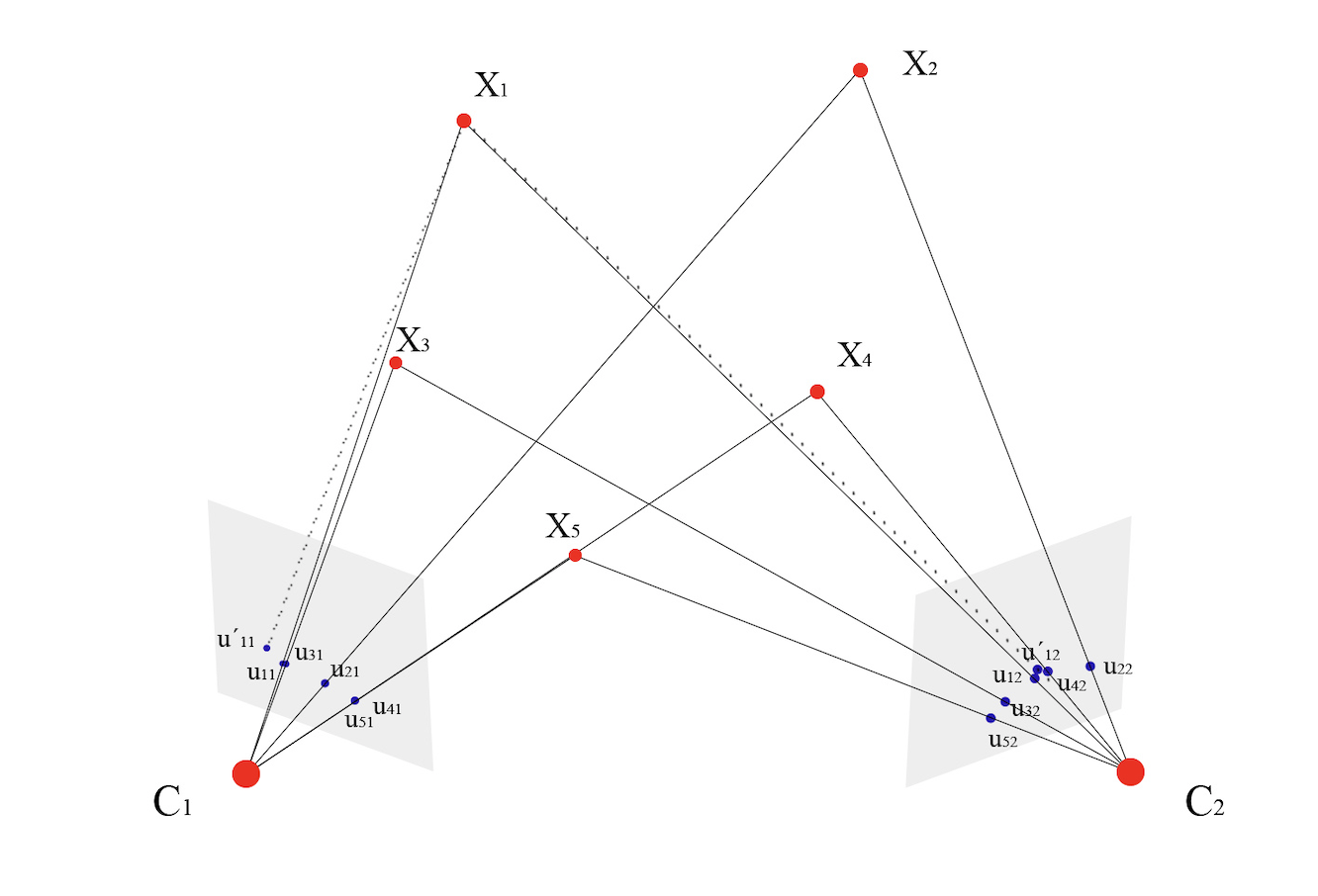}
\caption{A schematic representation of the Bundle Adjustment problem under consideration, involving 2 cameras and few points.}
\label{fig:bundle}
\end{figure} 
 
 Fig. \ref{fig:bundle} shows the problem studied with 5 out of 10 the points and 2 cameras.
 In Fig. \ref{fig:bundle} the $u_{ij}$ are the observation keypoints in the image plane, with $i$ the point label and $j$ the image label; $X_i$ are the points in the actual 3D frame; $C_j$ are the cameras, which are at different angles on different positions; finally, $u'_{ij}$ are the re-projected 2D points.  The solid lines represent projection lines and the dotted lines are representing the re-projection lines \cite{triggs1999,hartley2003}.
 Given this, the cost function can be calculated by the difference between the image keypoints and the reprojected points. One can define the residuals in several ways according to different metrics. In this work we use the root mean square reprojection error,
 \begin{equation}
     \label{eq11}
 S = \sum_{i,j} r_{ij}(\theta) = \sum^m_{j=1} \sum^n_{i=1} \sqrt{(\vec{u_{ij}} - \vec{u'_{ij} (\theta)})^2} \;,
 \end{equation}
where $n=10$ is the number of total points, and $m=2$ the number of total cameras, and $\theta$ the free parameters to be found in the minimization procedure.



The LMA is an extension of the Gauss-Newton and gradient descent methods. The assumption behind this algorithm is that the residual after a small step can be calculated in the linear approximation as,
\begin{equation}
\label{mainequation}
  r(\theta+\delta \theta)\approx r_m(\theta)+J_{m\mu}\delta \theta^{\mu}  
\end{equation}
$J$ is the Jacobian matrix defined by $J_{m\mu}=\dv{r_m}{\theta_\mu}$.
In the adjusted LMA used in this work, the step is defined with two damping parameters as,
\begin{equation}
\label{eq:lma}
    \delta\theta = -(J^TJ + \lambda_1 D^TD + \lambda_2)^{-1}J^Tr,
\end{equation}
where $D^TD$ is a diagonal matrix, which is determined by the relative scaling of the points and cameras and $\lambda_1$ is a damping parameter, which is being adjusted each iteration within the algorithm, and $\lambda_2$ is an additional damping parameter to update the solution as linear combination of the pure LMA step and the previous step \cite{triggs1999,hartley2003,transtrum2012improvements}. The inversion of the linear problem, that is the Schur complement of the linearized Hessian matrix $-(J^TJ + \lambda_1 D^TD + \lambda_2)$ \cite{triggs1999}, is the computational bottleneck of the LMA optimization algorithm and of the bundle adjustment procedure. 

Now it is possible to consider the application of a linear solver to find a solution to Eq. (\ref{eq:lma}) and therefore finding the step size of the LMA. The linearized Hessian is the matrix and $J^{T}r$ the vector of the linear system. Consequently, it is possible to apply either a classical linear solver or a quantum linear solver, such as the HHL algorithm.

\subsection{HHL}

HHL is one of the most promising algorithms to solve linear systems of equations. It is efficient both in the need for qubit memory and the scaling of the solution to the linear problem and inversion  \cite{hhl2009}. 

In the following, we will use the Dirac or "ket" notation to indicate quantum mechanical qubit operations and distinguish it from classical algorithmic steps. $\ket{x}$ is called a ket and is a vector in the Hilbert space, while $\bra{x}$ is a vector in its dual space called bra. Therefore the inner product between these two vectors, represented as $\bra{x'} \ket{x}$ is an element of the field. That is, a complex number. The norm of a quantum state vector is 1. For simplicity, we recount that bra and ket can be represented as horizontal and vertical vectors respectively in a familiar vector space over the real field. In this case, the inner product $\mathbf{x} \cdot \mathbf{x'}$ results in real numbers. The operators are matrices defined in this space. $\hat H$ is called the Hamiltonian, and is an operator over the vector space. In particular, the Hamiltonian describes the configuration and time evolution of the system. 
Considering an operator, which is represented by a square $N\times N$ (Hermitian) matrix $A$, the linear problem is defined as $|x\rangle:=A^{-1} |b \rangle$.

The solution of this problem can be obtained by considering the eigenvalues and eigenvectors of $A$ and representing $|x\rangle$ and $|b\rangle$ in the basis of the eigenvectors of $A$. In order to do so efficiently, the operator $A$ needs to be decomposed, together with the component-wise estimation of the eigenvectors. This is done through the Trotter--Suzuki method and the Quantum Phase Estimation (QPE) \cite{babbush2015chemical,dervovic2018quantum}. For clarity, an elementary example is provided in the 
Appendix B. 

\subsubsection{Trotter-Suzuki method}

The Trotter--Suzuki method is employed to apply operators to quantum states by approximating the application of a complicated operator by a sequence of simple ones and measuring the result in discretized time--steps. This is usually applied to the Hamiltonian, which defines the evolution a physical state in time. Instead of having a physical Hamiltonian, we have the matrix $A$ defining the linear problem.
Our operator $A$ is used as Hamiltonian to evolve the state in time. This operator needs to be decomposed, so we can write it in terms of quantum gates that act on a small number of qubits (usually either one or two qubit gates) \cite{nielsen_chuang_2010}. Any linear operator $A$ can be written as linear combination of matrices $\sum_j^n A_j$. However, $A_j$ terms do not commute with each other in general. Hence, the time evolution that looks like $e^{-i\sum_j^n A_j t}$, can be decomposed in product of the evolutions of the single terms,
\begin{equation}
    e^{-i t \sum_{j=1}^n A_j} = \prod_{j = 1}^n e^{-i A_j t} +\mathcal{O}(n^2t^2),
\end{equation}
which however is not exact because of the non--commutation. If the time $t \ll 1$ this linearization is precise. However, a large amount of quantum gates are needed for evolving in small time-steps in order to measure the outcome of operator application several times.
Nonetheless, increasing the number of gates reduces the success probability. To raise the precision it is possible to split our decomposition into $r$ steps,
\begin{equation}
    e^{-i\sum_{j=1}^n A_j t} = \left(\prod_{j = 1}^n e^{-iA_j\frac{t}{r}}\right)^r +\mathcal{O}(n^2t^2/r),
\end{equation}
which results in an improved precision and time scaling. This can be built on, which is called the second order Trotter-Suzuki method. The idea is to cancel error terms by constructing a sequence of operators, such as,
\begin{align}
\label{times}
e^{-iAt} = \left(\prod_{j=1}^ne^{-iA_j\frac{t}{2r}}\prod_{j=n}^1e^{-iA_j\frac{t}{2r}}\right)^r + \mathcal{O}(n^3t^3/r^2).
\end{align}
The second order Trotter--Suzuki method is used in this work to calculate the state evolution of the operator representing the linear system \cite{DERAEDT19871}. The operator $A$ is decomposed in basic one and two qubits gates $\sum_j A_j$ in order to solve the linear system (cf. Table \ref{table:gates} for a list of gates used for the present implementation). To be noted that this approximation is not exact.



\subsubsection{QPE}
Quantum phase estimation is a powerful tool in quantum information. Its general purpose is to find the eigenvalues corresponding to eigenvectors. 
In this problem, QPE is used to decompose the vector $\ket{b}$ in the linear equation into the eigenbasis of $A$ and find respective eigenvalues. This can be done by initializing our qubit states $\ket{b}$ and an ancilla register with three qubits, where we set each qubit into a superposition of $(\ket{0}+\ket{1})/\sqrt{2}$. Then we use a control $U$ gate, which applies the operation $U=e^{-iAt}$ if the control qubit is $\ket{1}$. The control is on the ancilla register, meaning that only in the $\ket{1}$ ancilla state we apply the unitary operation on the target state, which looks like,
\begin{align}
\label{QPEE}
    \begin{split}
    \psi_2 = \frac{1}{2^{3/2}}\left( \ket{0}+e^{2\pi i \theta 2^{2}}\ket{1}\right)\otimes \left(\ket{0}+ e^{2\pi i \theta 2^{1}} \ket{1}\right)\\
    \otimes \left( \ket{0}+e^{2\pi i \theta 2^{0}}\ket{1}\right)\otimes \ket{b},
    \end{split}
\end{align}
where we split $U$ into $U^{2^j}$ to result in a binary representation of the action on the ancilla. In general,
\begin{align}
    U^{2^{j}}\ket{b} = U^{2^{j-1}} U\ket{b}=e^{2\pi i\theta 2^j} \ket{b}.
\end{align}
Finally, we can use a Inverse Fourier Transform to change our Fourier basis to computational basis loading our eigenvalues into the ancilla register in binary representation \cite{Qiskit-Textbook}.

\subsection{Implementation}
A simple test pipeline for bundle adjustment was developed in python, following a structure inspired by Ceres \cite{ceres-solver}. The objective of this routine is to have a simple and pedagogical way to test different optimization procedures. This pipeline includes jet evolution to calculate analytical derivatives. The initial state, that is points $X_i$ and camera positions and orientations $C_j$ in Fig. \ref{fig:bundle}, are defined arbitrarily in coordinates and quaternions respectively. The exact solution is backward propagated to the image planes $u_{ij}$ to generate observation keypoints. Finally, uniform random noise is added to keypoints and initial camera guesses. The non normal noise makes the cost landscape challenging to navigate. 

In this pipeline we implemented both the classical adjusted LMA and the quantum LMA in python. Still inspired by \cite{ceres-solver}, the damping parameters of Eq. (\ref{eq:lma}) are updated as follows. After initialization, $\lambda_1$ is updated at each iteration depending on the average value of the dot product of the gradient and the previous iteration's step, $\Omega$. $\lambda_1$ is then updated at each iteration by multiplying by $\lambda^+$, either if $\Omega$ is positive, or if the difference in the squared residual is smaller than $-\frac{1}{4}\Omega$. This choice indicates an increase or a very small reduction of the residual, or a change of direction in the optimization. Otherwise, if the squared residual difference is bigger than $-\frac{1}{2}\Omega$, so if the residual decreases sensibly and the direction of the optimization step doesn't change, we multiply $\lambda_1$ by $\lambda^-$.

\begin{table}[h!]
\centering
\begin{tabular}{ |c|c|c| } 
 \hline
  & Setup 1& Setup 2\\ 
 \hline
 \hline
 $\lambda_1$ & 0.01 & 0.0001 \\\hline
 $\lambda_2$ & 0.01 & 0.0001 \\\hline
 $\lambda^+$ & 1.5  & 1.1    \\\hline
 $\lambda^-$ & 0.7  & 0.9    \\\hline
\end{tabular}
\caption{Two different setups with each different damping parameters and their updating strength. Damping parameters and update multipliers used in sparse LMA and QLMA.}
\label{table:parameters}
\end{table}

The classical Levenberg Marquardt makes use of Yale sparse format, which allows fast row access. Then the linear system in Eq. (\ref{mainequation}) was solved using the method \texttt{\;spsolve} from the python package \texttt{\;scipy.sparse.linalg} \cite{spsolver}. 

The quantum implementation of the Levenberg-Marquardt algorithm includes the same setup as the classic LMA, the difference lays exclusively in the solving of the linear problem via the HHL algorithm with the implementations of Trotter--Suzuki evolution and Quantum phase estimations described above. 50 time slices $r$ were chosen for the Trotter--Suzuki evolution as from Eq. (\ref{times}). $A$ and vector $b$ were written in dense format instead of sparse. Furthermore, the matrix and vector have to be expanded to size $2^n$ to be represented by qubits. 
All the results shown have been obtained using HHL algorithm implemented in python package qiskit aqua version 0.8.2, with qiskit's general version being version 0.23.6. However, previous versions have also been used in development.

Each of the test run was executed for 3 days on a computing cluster node with dual Xeon E5-2650 v3 processors.
64GB of memory are necessary to simulate the quantum computer qubits necessary to run the HHL algorithm for the present problem with qiskit.

\section{Results}
\subsection{Results of simulation}

\begin{figure}[h!]
\centering
\includegraphics[width = 7.95cm]{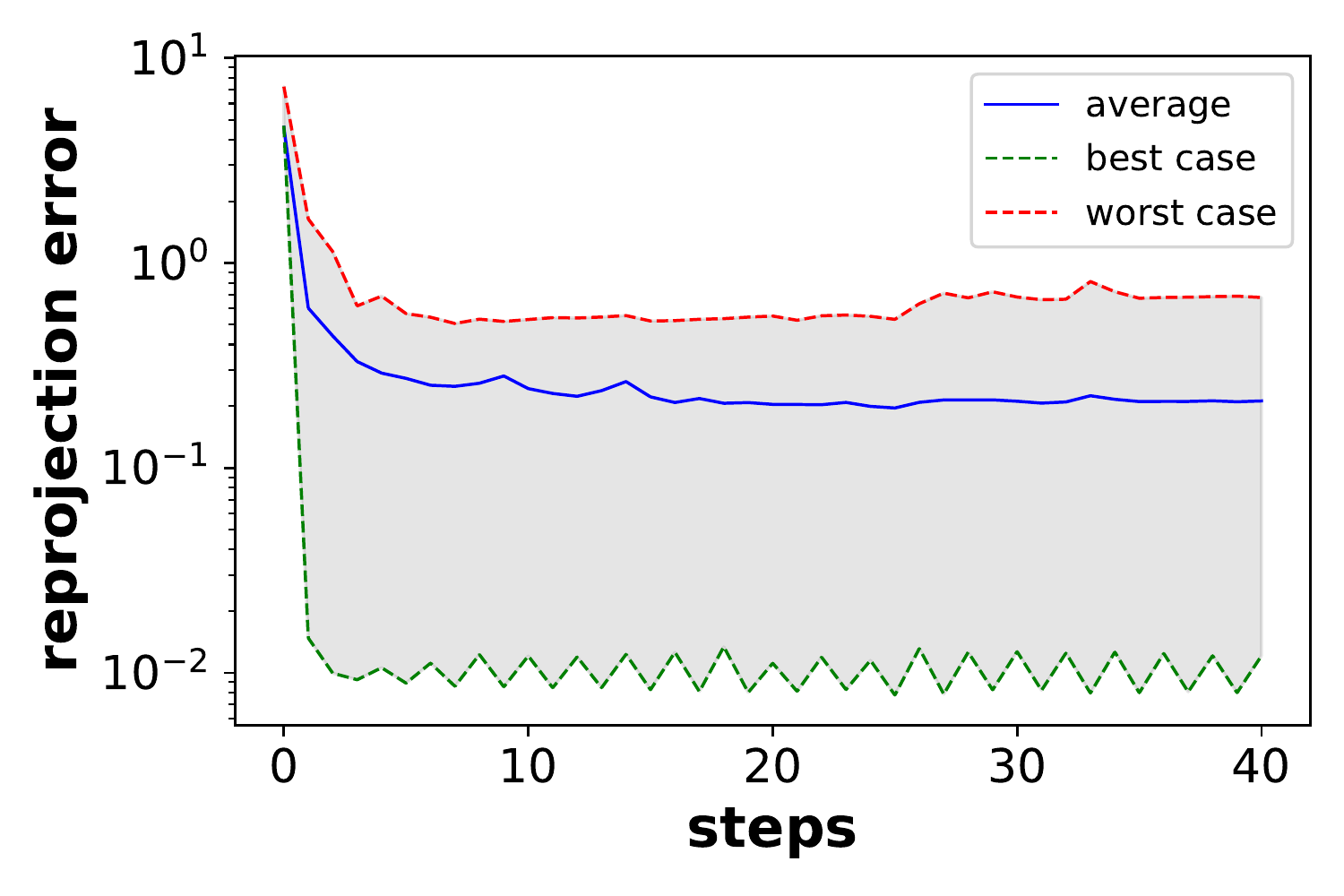}
\caption{Total reprojection error in function of the QLMA optimization step, using damping parameters Setup 1 from Table \ref{table:parameters} for 9 problems. The blue line represents the average cost of the 9 different problems, the red the worst case and the green line represents the best case, defined by the best and worse costs at the last iteration.}
\label{hhl}
\end{figure}

\begin{figure}[h!]
\centering
\includegraphics[width = 7.95cm]{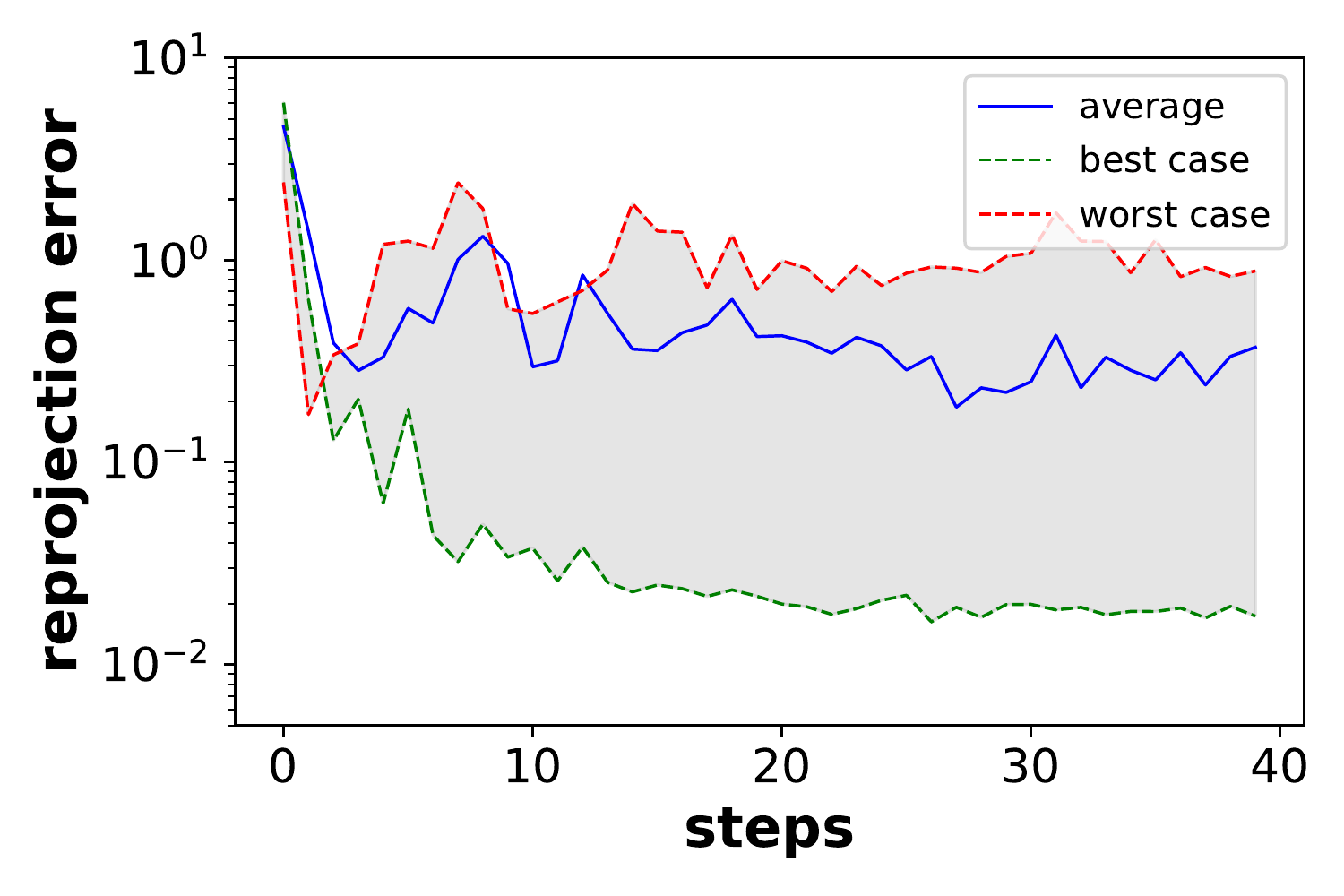}
\caption{Total reprojection error in function of the sparse LMA optimization step, using damping parameters Setup 1 from Table \ref{table:parameters} for 9 problems. The blue line represents the average cost of the 9 different problems; the red the worst case and the green line represents the best case, defined by the best and worse costs at the last iteration.}
\label{lma1}
\end{figure}

\begin{figure}[h!]
\centering
\includegraphics[width = 7.95cm]{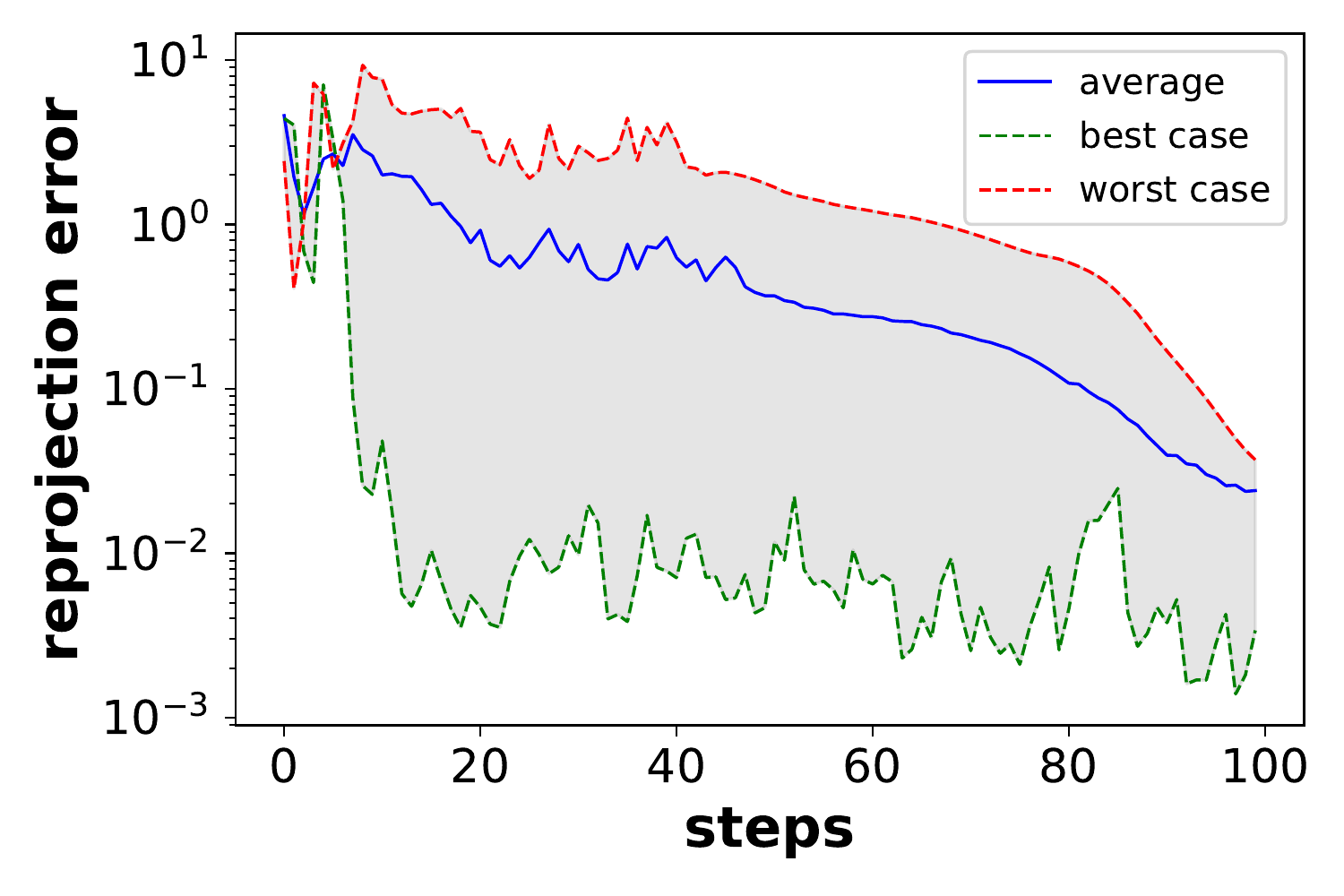}
\caption{Total reprojection error in function of the sparse LMA optimization step, using damping parameters Setup 2 from Table \ref{table:parameters} for 9 problems. The blue line represents the average cost of the 9 different problems; the red the worst case and the green line represents the best case, defined by the best and worse costs at the last iteration.}
\label{lma2}
\end{figure}

\begin{figure}[h!]
\centering
\includegraphics[width = 7.95cm]{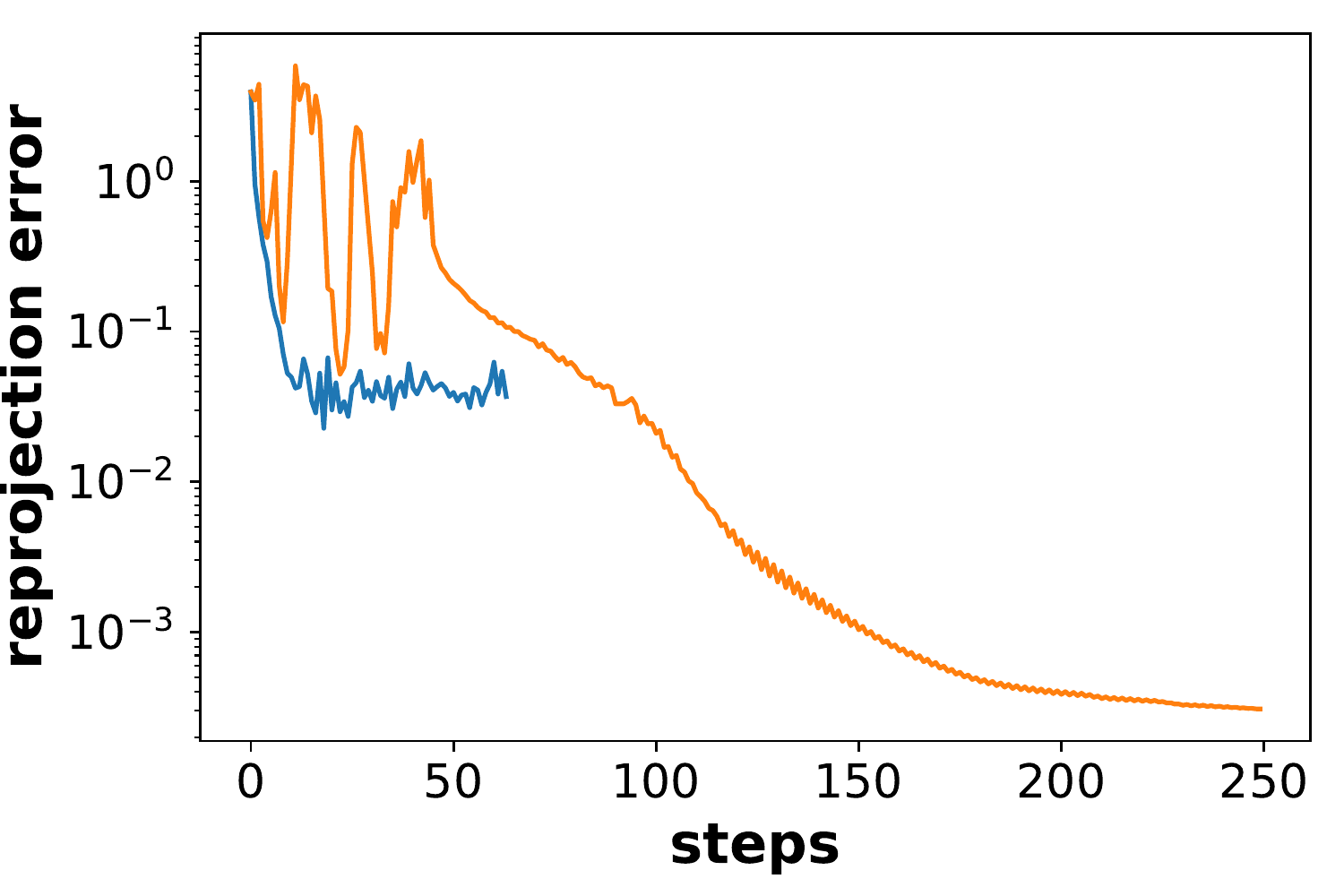}
\caption{Comparison of the reprojection error for LMA (orange line) and QLMA (blue line) in problem 8 with damping parameters from Setup 2.}
\label{hhllma3}
\end{figure}

We solve 9 different setups with the QLMA and the classical LMA, which can be seen in Figures \ref{hhl} to \ref{hhllma3}. The problems include 10 points, which are initialized at random with different seeds, and two cameras. After initialization we add uniform noise. The noise is 25\% of the extremes of the position of the points (points are set at random between -2 and 2, the noise moves each of them between -0.5 to 0.5 from the assigned position) and 50\% of the position and orientation of cameras. The uniform and abundant noise, together with the fact that only two cameras are provided, make the optimization landscape more challenging than an average bundle adjustment problem with adequate amount of data and preconditioning.

For this problem, the matrix to be inverted is a square matrix of size 12, that is the number of parameters of the Schur complement. This results in an expanded size of 16 columns and rows, to represent it in qubits. This is further expanded to 32, for creating an Hermitian matrix. This results in a qubit number of 5 qubits to encode the data. Further 5 qubits are then used for the ancilla registers, of which 3 for the QPE algorithm.

In Figures \ref{hhl}, \ref{lma1} and \ref{lma2} the blue line represents the average cost of the 9 different problems. It can clearly be seen that in Fig. \ref{lma1} the damping parameters of the first setup are too large for reaching convergence in this difficult and not tightly constrained landscape with the sparse LMA. In Fig. \ref{lma2} a better convergent behaviour already in the first 100 iterations can be reached using the second setup of damping parameters. However, it can be seen in Fig. \ref{hhl} that the QLMA provides very smooth optimization even within the difficult landscape and harsh damping parameters of setup 1. We could only run one calculation with the second damping parameter setup, where again QLMA outperforms LMA in the first iterations, cf. Fig. \ref{hhllma3}. The reprojection error reaches the minimum within the first 10 iterations. The cost is not optimized further for several reasons: the setup of damping parameters, and the approximate nature of Trotter--Suzuki and QPE. Surprisingly, the smooth behaviour and rapid convergence of QLMA might come precisely by the sophisticated linearization approximations adopted in the HHL resolution. Note that some of these linearization procedures could be applied in classical computing as well.

Another fact to highlight is that the QLMA and LMA perform differently well in different problems. When the landscape is more challenging and fraught with local minima and steep derivatives, the LMA with these parameters often overshoots, generating noise in the cost graph, or gets stuck in the wrong minimum, while QLMA quickly converges to an acceptable cost. When the landscape is amenable and smooth, LMA quickly converges with the same number of iterations (cf. 
Appendix A).


\subsection{Probability of success and limitations}
On the computing cluster used (cf. Acknowledgments), the computing time limit compatible with our queue time was 3 days per run. This limited us to approximately 40 iterations each run. Nonetheless, for all problems under considerations the quantum algorithm reached an approximate convergence within the first 40 iterations. 
The time scaling of HHL depends on several factors, including the sparseness of the matrix, choice of ancilla registers, and approximation degree of the solution obtained, both in terms of precision and in number of components extracted from the linear problem. In particular, note that to obtain the full solution is necessary to run HHL a number of times directly proportional to the dimension of the problem. However, this is not strictly necessary to obtain an approximate solution within the basis of eigenstates of $A$, corresponding to an efficient truncation over the principal components. The number of measures necessary to an effective optimization is not known a--priori.
Furthermore, several specific implementations and variations try to improve on the original HHL and provide full or approximate solution with even greater efficiency \cite{dervovic2018quantum,lee2019hybrid}. For a practical use of this algorithm and analysis of its performance, a quantum hardware with orders of magnitude improvement in quantity and quality of qubits would be necessary, which to the current date are still out of reach.

From the complexity point of view, time complexity of this algorithm for a given numerical precision sits between $O(\log(m))$ and $O(m^{3/2}\log(m))$, with $m$ the number of parameters in the Schur complement, depending on the previously mentioned factors. The space complexity of the HHL, is more clearly $O(\log(m))$, only $2n$ qubits are necessary to solve the $2^n \times 2^n$ problem, instead of $O(m^2)$ of classical algorithms. The complexity of classical LMA is $O(m^3)$ in time and $O(m^2)$ in space \cite{hhl2009,dervovic2018quantum}. Therefore, once quantum computing becomes viable and scales, the approach of this problem could rapidly be feasible in large scale. This toy problem required 10 qubits, of which only 5 to store data. However, with only 51 ideal qubits it would be possible to represent a Schur complement for 33 million parameters. Therefore, it would rapidly exceed the largest Bundle Adjustment problems solved \cite{zhang2017distributed}. Unfortunately, this supposes unrealistically perfect qubits. In practical implementations the need for additional ancilla registers and noise correction increases the physical qubits requirements dramatically (as a title of example, cf. literature on Shor's factorization with physical qubits \cite{gidney2021factor}).

To date, the most ambitous goals are set by IBM to create a 1 million qubit quantum computer by 2030. The necessary changes in the previously mentioned aspects very likely make further investigation on noise behaviour and realistic physical implementation and scaling of QLMA indeed premature.

We can direct our attention further to a possible implementation of our problems to a hardware quantum computer. There is constant progress on noise mitigation and noise control on quantum hardware and software. Recently in an experimental configuration the error rate with quantum error mitigation of single gates was contained down to the magnitude of $10^{-5}$ or two qubit gates were measured with error rates of $5\cdot 10^{-3}$ \cite{zhang2020error,garion2021experimental}.
For IBM-Q public quantum computers, the nominal error rate of gates are of magnitude of $10^{-2}$ for two qubit gates and $10^{-3}$ for one qubit gates \cite{zulehner2020introducing}. Finally, after the gates operations, results must be measured. For example, the Melbourne quantum computer has an average measurement error rate of 0.0812 \cite{tannu2019mitigating}.

Realistic noise models are more complicated than a simple number associated to gates operation and measurements. Different qubits and gates have different noise on actual quantum hardware. Nevertheless, for a simple estimate one can consider these noise rates and multiply by the number and type of gates. 

In the quantum simulation done by us, we construct a gate setup with gates as can be seen in Table \ref{table:gates}. As there are 60 one qubit gates and 118 two qubit gates and have a width of 10 qubits, one can find the mean average success rate of approximately $\approx 21\%$ for a realistic IBM-Q \cite{zulehner2020introducing} and $\approx 56\%$ for the best present experimental results \cite{zhang2020error,garion2021experimental} (note that presently the best results are obtained on different setups). This completely excludes any kind of coherence errors from the quantum computer.
There has been a discussion raised regarding the coherence error of a circuit with a depth of 300 and it was estimated to lower the chance of success from 50\% to 10\% on the IBM-Q quantum computer Melbourne \cite{IBMQ}. Using those approximations, it can be believed that in our case the success rate for one useful iteration could be lower than 10\%. If we then would want to run 10 iterations without failures the success probability is at $10^{-10}$.

\begin{table}[h!]
\centering
\begin{tabular}{ |c|c| } 
 \hline
Gate Type & Number of Gates\\ 
 \hline
 \hline
 X & 24 \\\hline
 U & 30 \\\hline
 H & 6  \\\hline \hline
 CU & 42\\\hline
 CX & 76\\\hline
\end{tabular}
\caption{Number of elementary one and two qubit gates in which the HHL algorithm is divided into in the case of 10 points and 2 cameras under consideration. Where the notation means X, Pauli X gate, U general rotation gate, H, Hadamard gate, CX the Controlled Not, and CU the Controlled rotation gates (cf. \cite{nielsen_chuang_2010,Qiskit-Textbook} for more information).}
\label{table:gates}
\end{table}

Another possibility for investigation could be including noise simulations of current quantum computer and study the effect of different kinds of error mitigations, such as the read-out mitigation or exponential extrapolation \cite{roggero2020preparation}. In this paper we did not include any noise simulations, which are a significant part in current quantum computing. 

\section{Conclusion}

The results of this paper indicate that the linearization techniques such as Trotter-Suzuki and QPE offer a more robust convergence in less iterations for difficult problems compared to a direct application of LMA. Some of these techniques could be studied as linear algebra methods, independently from their quantum computing origin.

Furthermore, this work investigates the potential application of quantum computers in the field of computer vision, for optimization problems, and bundle adjustment. Although, the realization of such an application is currently limited by various issues and necessary development in quantum hardware. 

A possible next step for the theoretical research would be to run the proposed algorithm with different problems and a higher number of ancilla qubits. The residual in our simulation does not drop below $10^{-2}$ even if the damping parameters are adjusted or a simpler problem is solved. We believe this is due to the fact that Trotter Suzuki and QPE approximations use at most 5 ancilla qubits in our case. Increasing the number of ancilla qubits allows for a higher precision in the time evolution and measurement, and therefore an investigation on the source of residual limit. 

Additionally, several possible quantum algorithms could be further investigated that could offer several advantages over our hybrid implementation. Including multiple quantum algorithms developed for noisy linear problems \cite{Grilo2019}, the Variational Quantum Eigensolver optimizer \cite{ferguson2021measurement} and rapid single--lookup numerical derivative estimators such as the Jordan algorithm \cite{jordan2005fast}. These approaches combined could help create a fully Quantum optimization algorithm for bundle adjustment. Such an implementation could lead to further speed ups and potentially overcoming several obstacles in optimization such as noisy data and local minima. Nonetheless, the number of gates and qubit requirement would arguably exceed the already steep requirements of HHL.

The current results presented in this paper indicate promising potential for future implementations of quantum algorithms within bundle adjustment. In the future, quantum computers with higher qubit number and improvement on noise could use and re-examine these results. 
Specific mathematical problems such as factorization are already being successfully tackled by quantum algorithms like the Shor algorithm \cite{shor1999polynomial}. However, no quantum algorithm has been demonstrated to solve a wide range of mathematical and computer scientific problems. 

The purpose of this study is to show that this Quantum Levenberg--Marquardt algorithm has the potential to tackle complex optimization problems. This despite that computational limitations force to construct an artificial bundle adjustment problem with very few parameters at the present moment. However, it can be seen that the proposed algorithm offers interesting advantages, such as an efficient scaling, and also smoother and faster convergence for this problem, without the need for preconditioning. 
With this in mind, it will very likely take still considerable time, stretching several years, for large scale quantum computer being able to practically solve problems like large reconstructions with bundle adjustment. 

\section*{Acknowledgments}
We thank Maurizio Monge for discussions and help concerning the classical Levenberg--Marquardt implementation. The computations were enabled by resources provided by LUNARC computing cluster. A.I. was supported by the Crafoord foundation and Swedish Research Council 2020-03721.

\appendix
\section{Additional results}

We present here the explicit results of the runs comparing LMA to the HHL. In Figs. 2-4 of the main text we see the performance of QLMA compared to the LMA with respect to an average across 9 different problems and the extremes of best and worse cases. That is, the problems that give highest or lowest cost at convergence of the algorithm. In Figure 5 of the main text, the comparison for one problem with setup 2 of damping parameters for both LMA and QLMA.

The quantum algorithm already converged mostly within the first 40 iterations, whereas the LMA did have a very hard time with the parameters of setup 1. When using setup 2 the LMA did considerably well, which can be seen in an example in Fig. 5 of the main text. Nonetheless, the HHL did show only modest improvements on convergence due to what we believe is the linear approximations employed in Trotter Suzuki expansion.

\begin{figure}[h!]
\centering
\includegraphics[width = 0.45\textwidth]{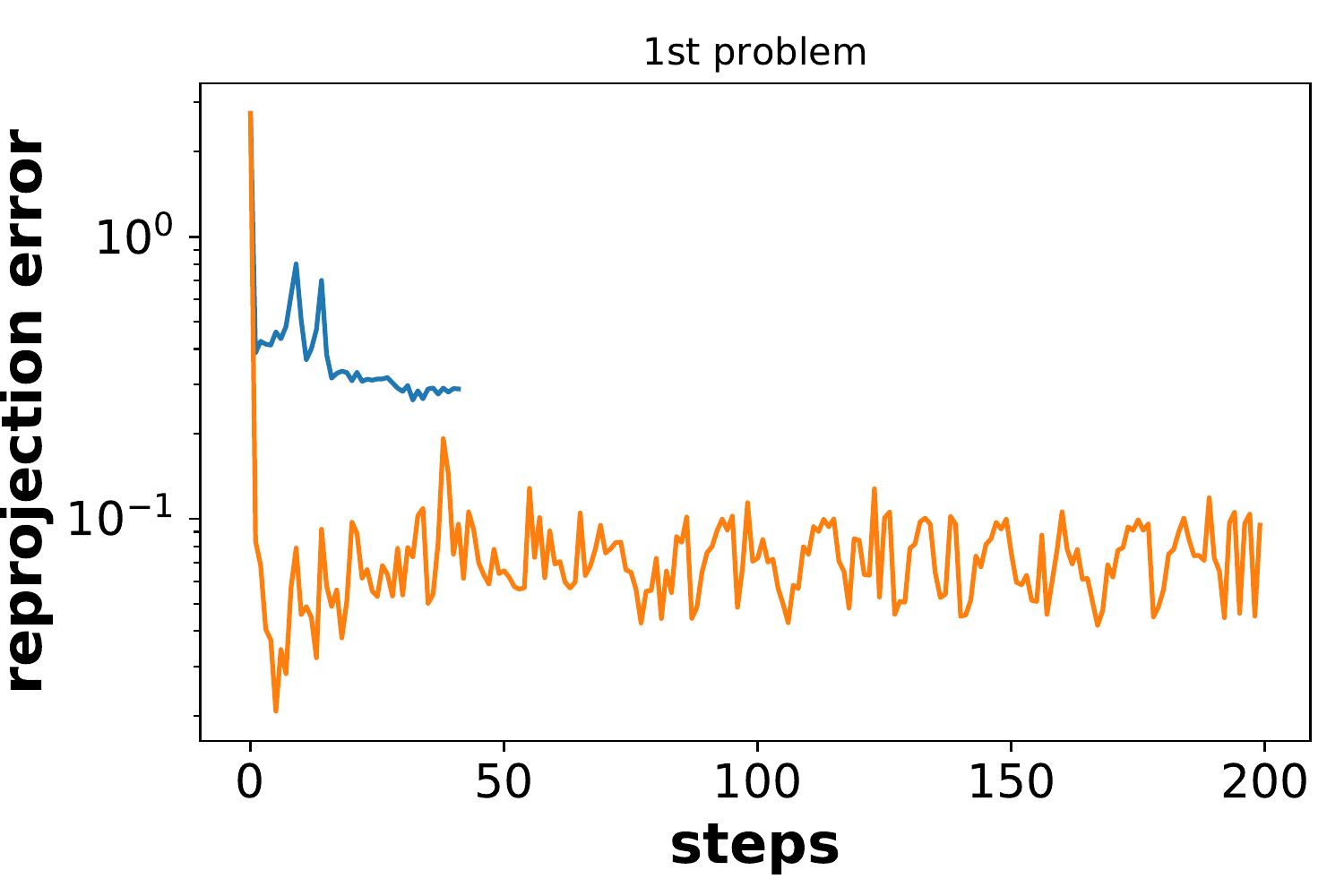}
\includegraphics[width = 0.45\textwidth]{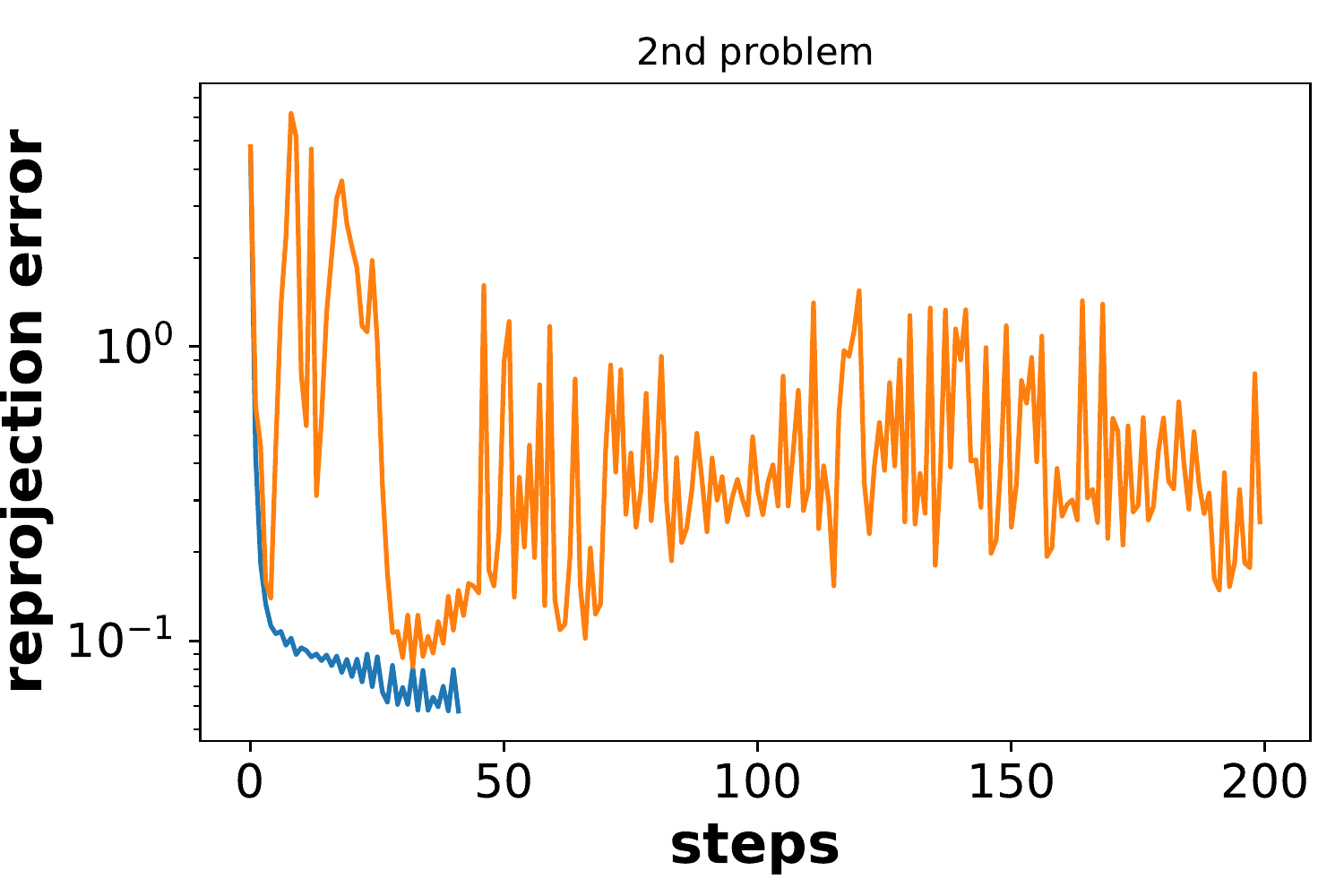}
\includegraphics[width = 0.45\textwidth]{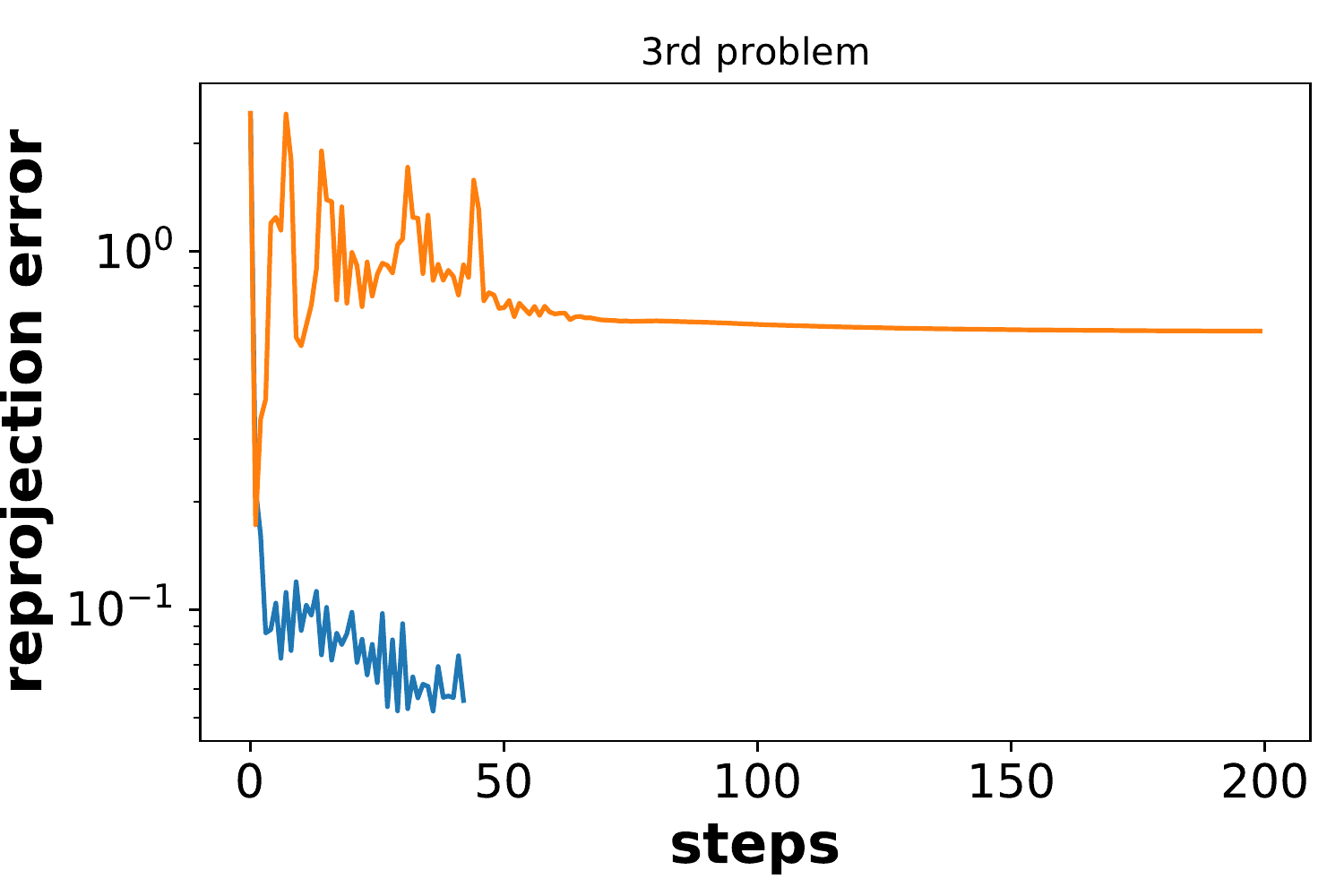}
\includegraphics[width = 0.45\textwidth]{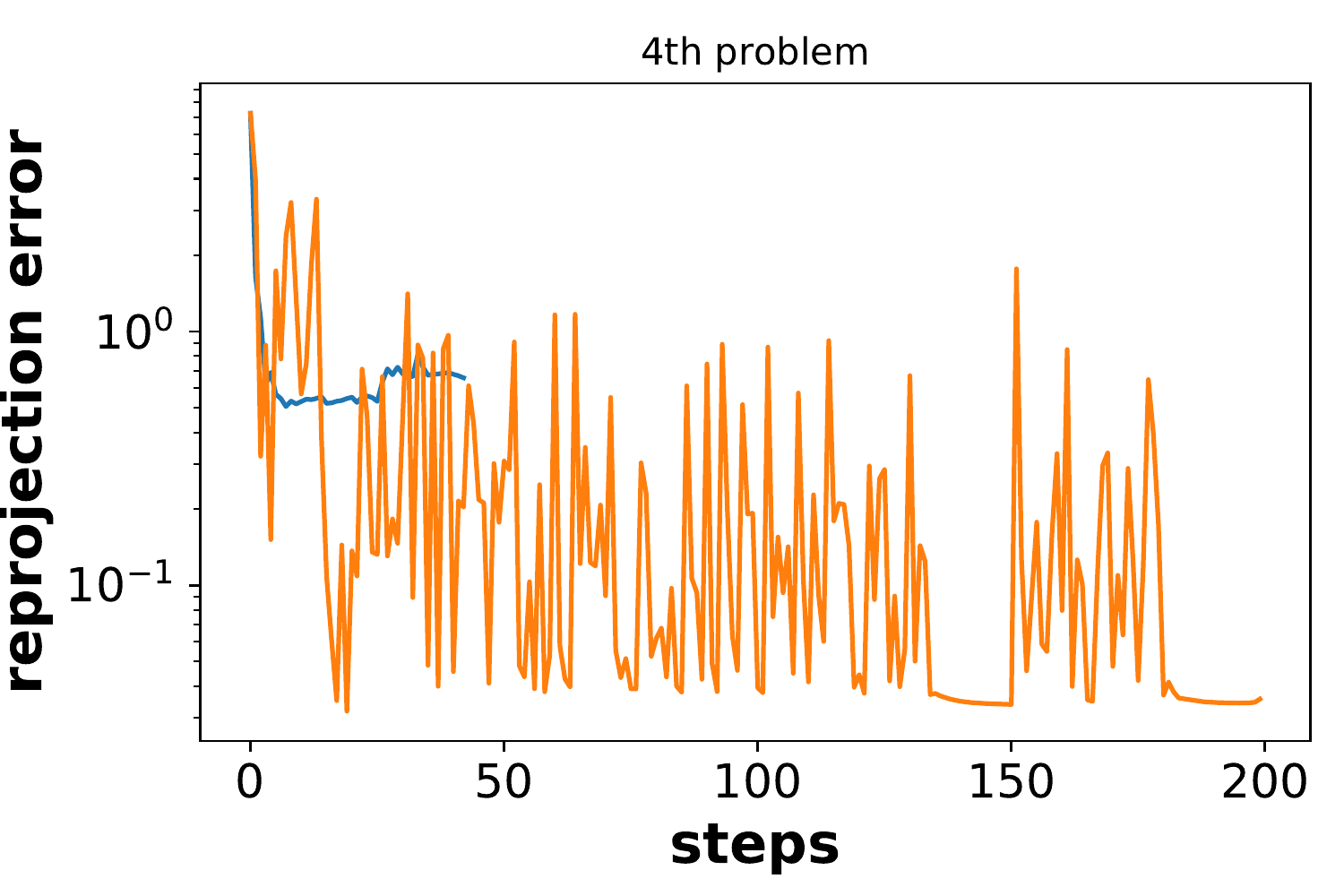}
\includegraphics[width = 0.45\textwidth]{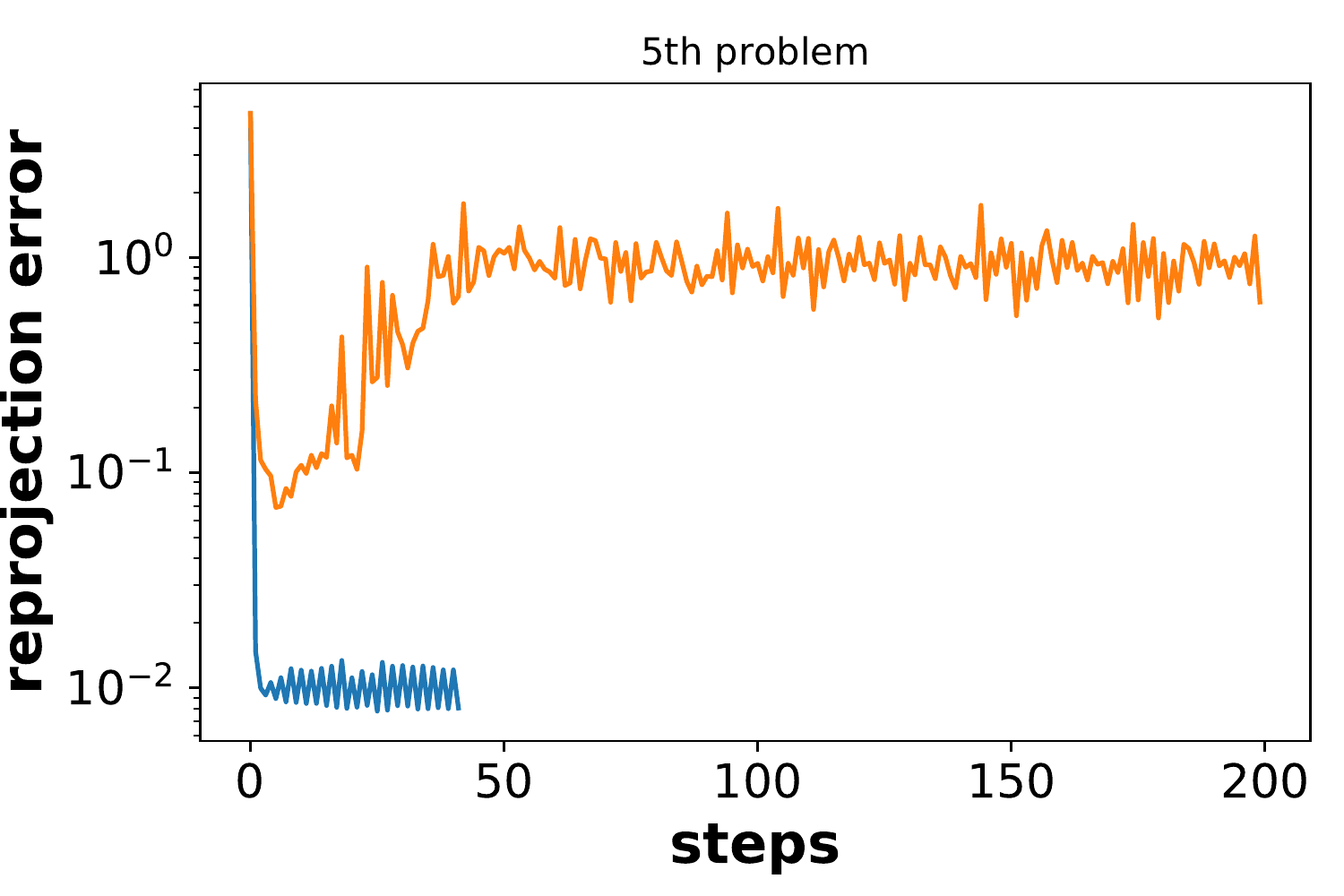}
\includegraphics[width = 0.45\textwidth]{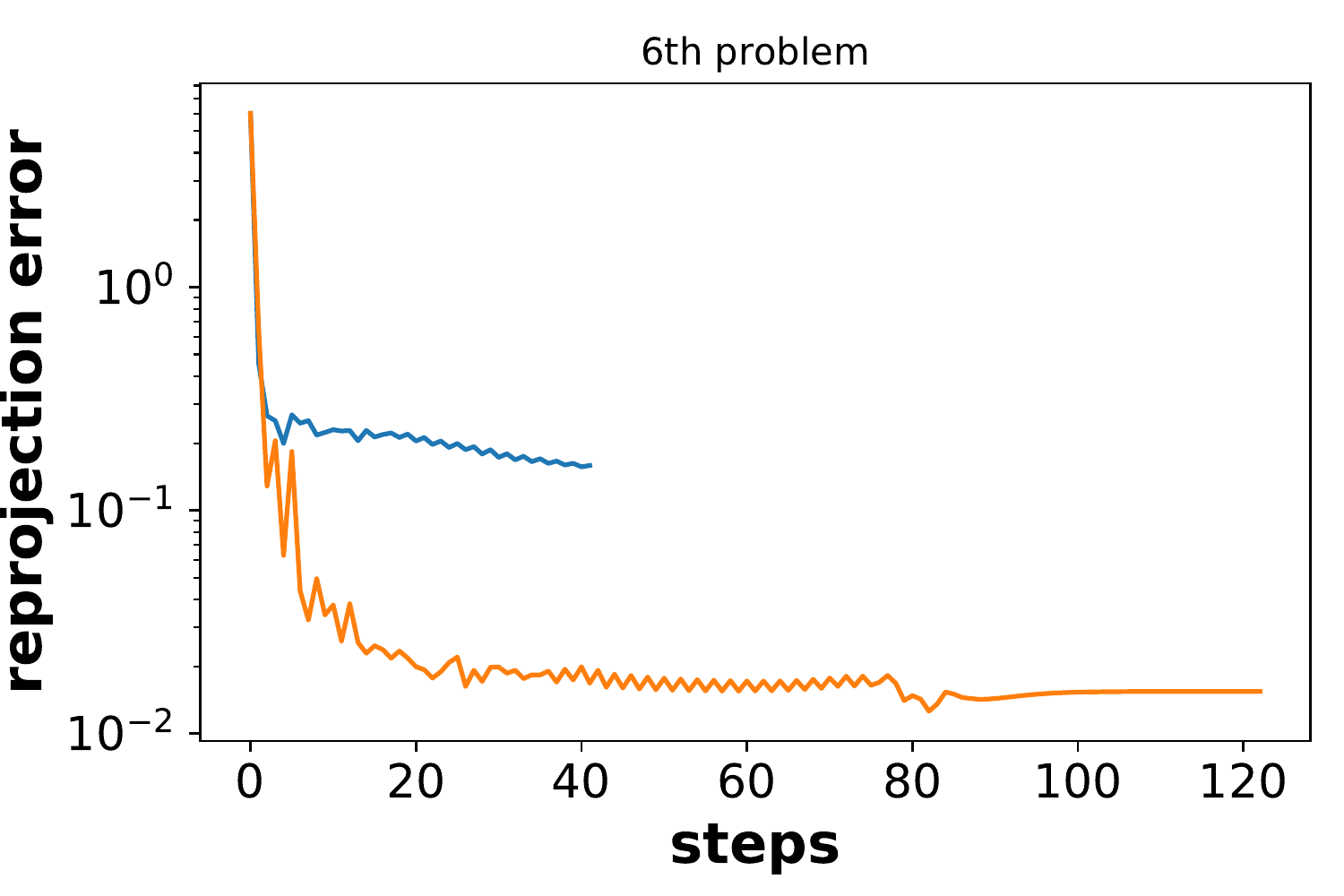}
\includegraphics[width = 0.45\textwidth]{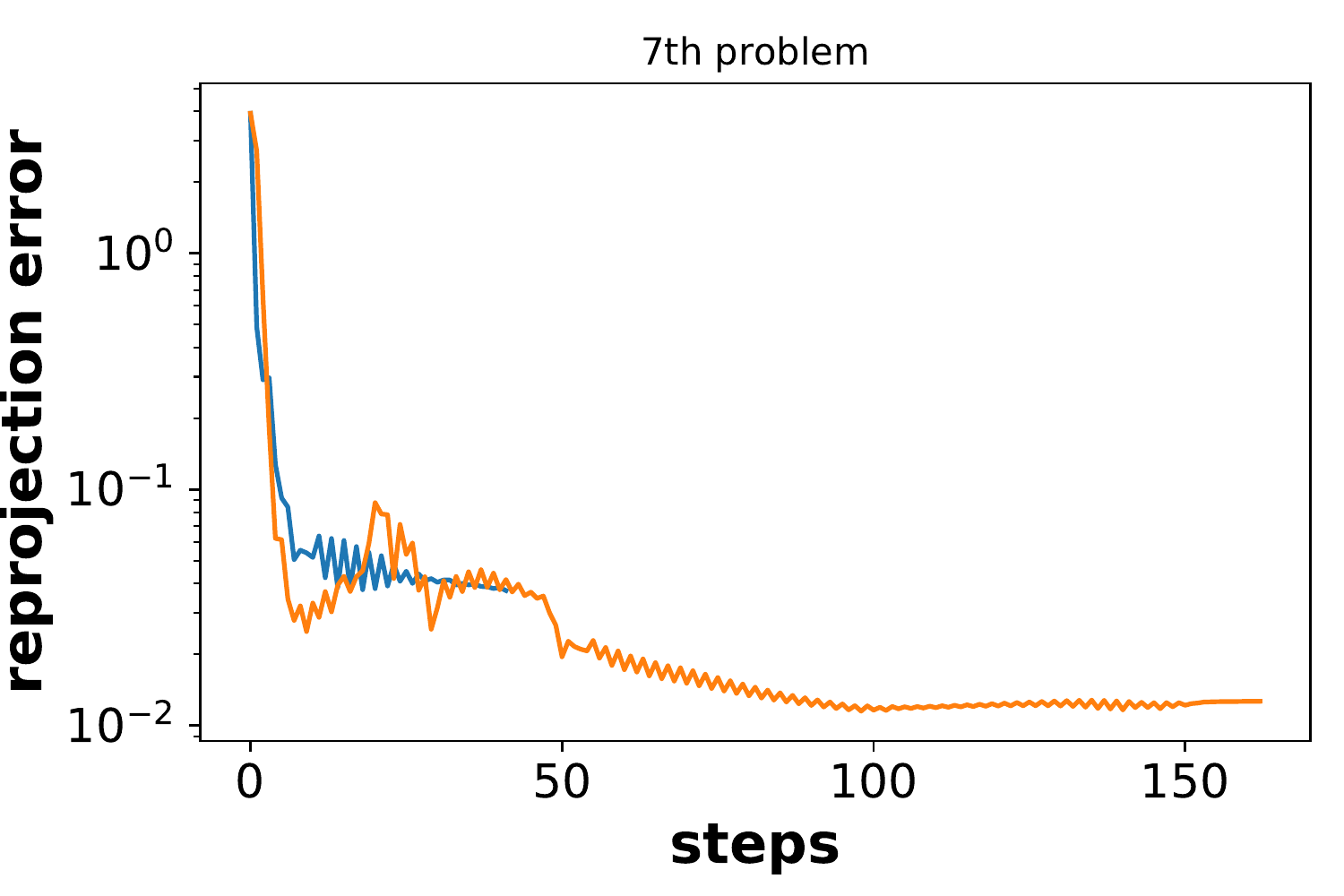}
\includegraphics[width = 0.45\textwidth]{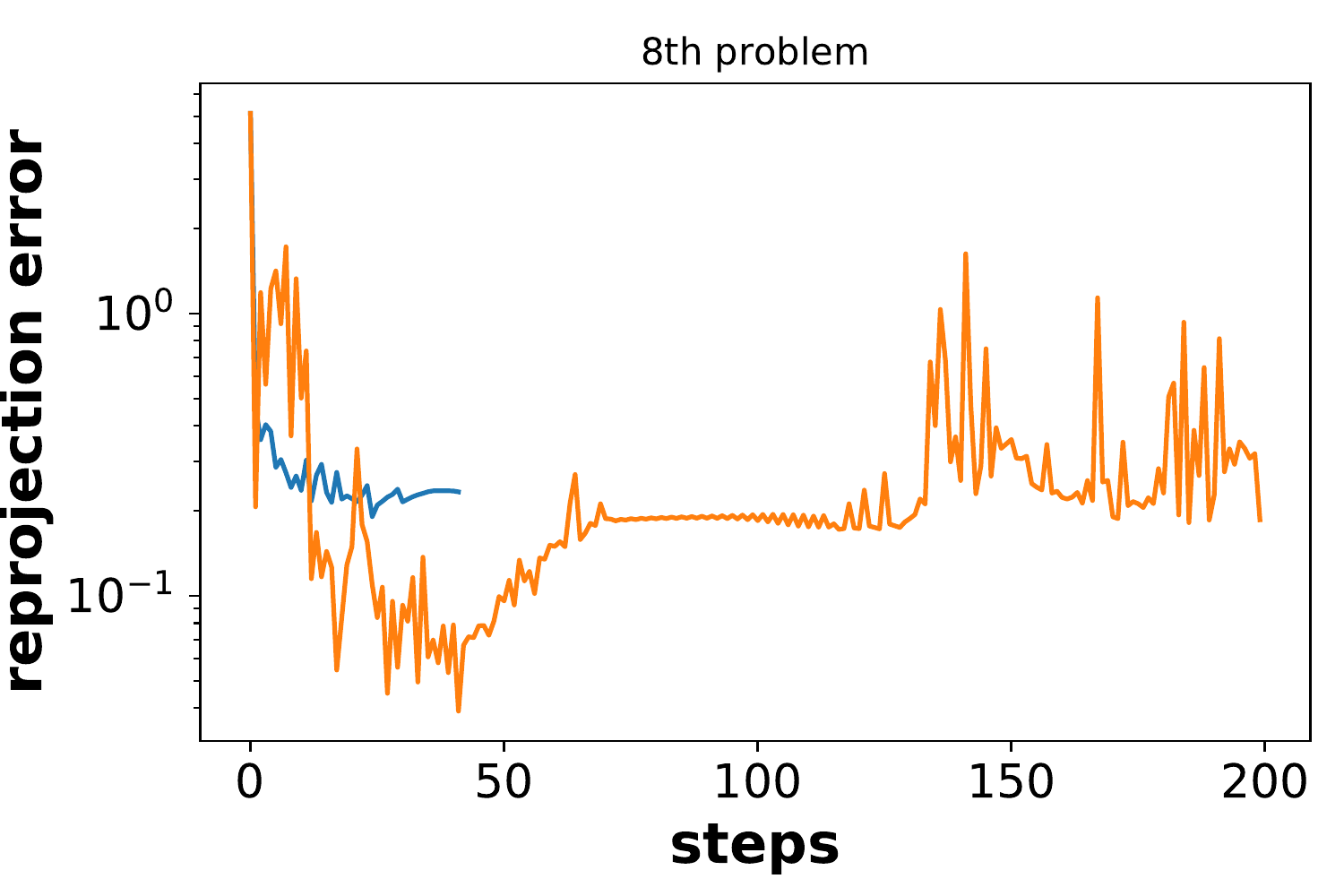}
\includegraphics[width = 0.45\textwidth]{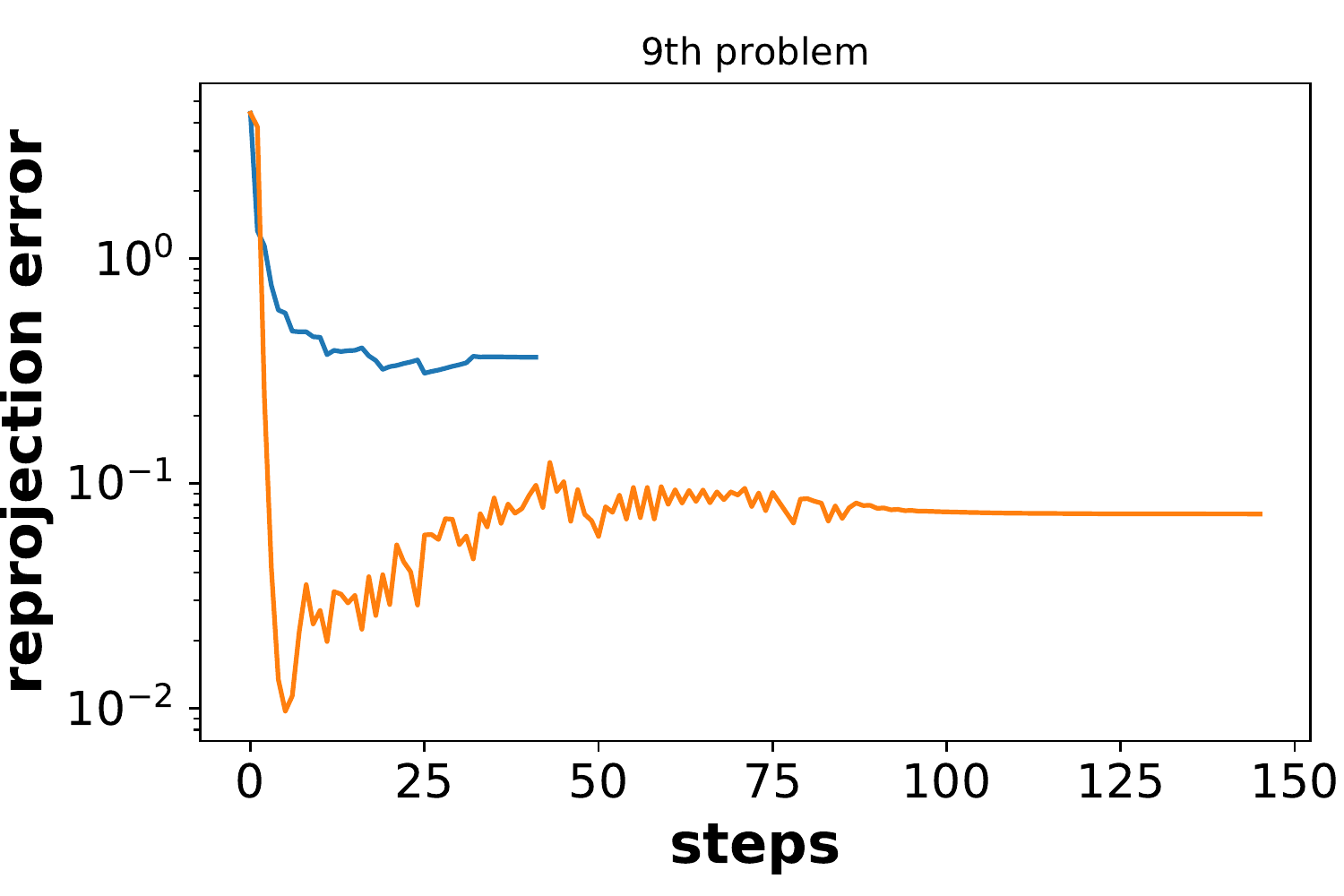}
\caption{Comparison of the sparse LMA to QLMA for the 9 random problems from setup 1 in Table 2 of the main text. The sparse LMA is denoted by the orange line and the QLMA by the blue line.}
\label{fig:6a}
\end{figure}

\begin{figure}[h!]
\centering
\includegraphics[width = 0.45\textwidth]{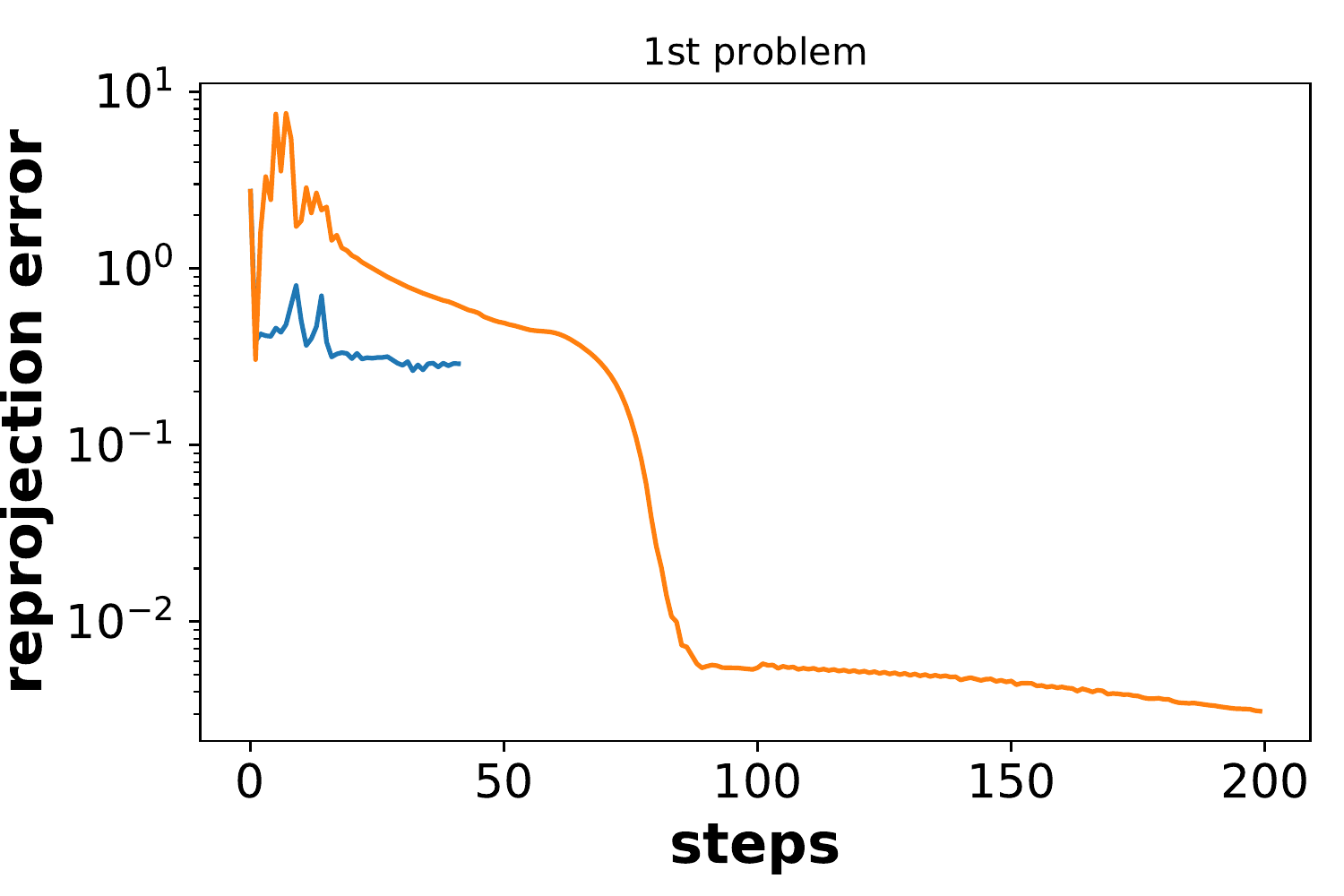}
\includegraphics[width = 0.45\textwidth]{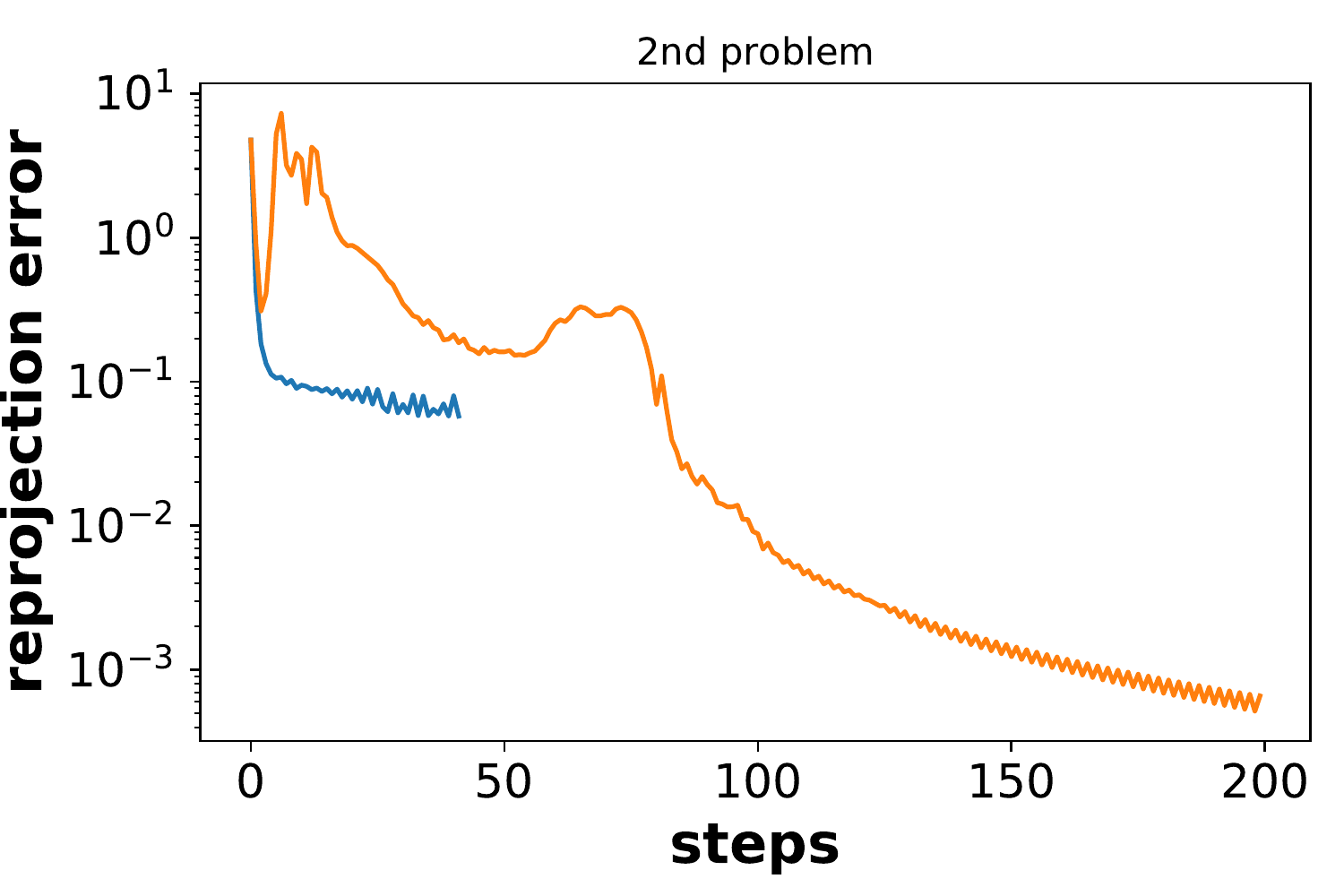}
\includegraphics[width = 0.45\textwidth]{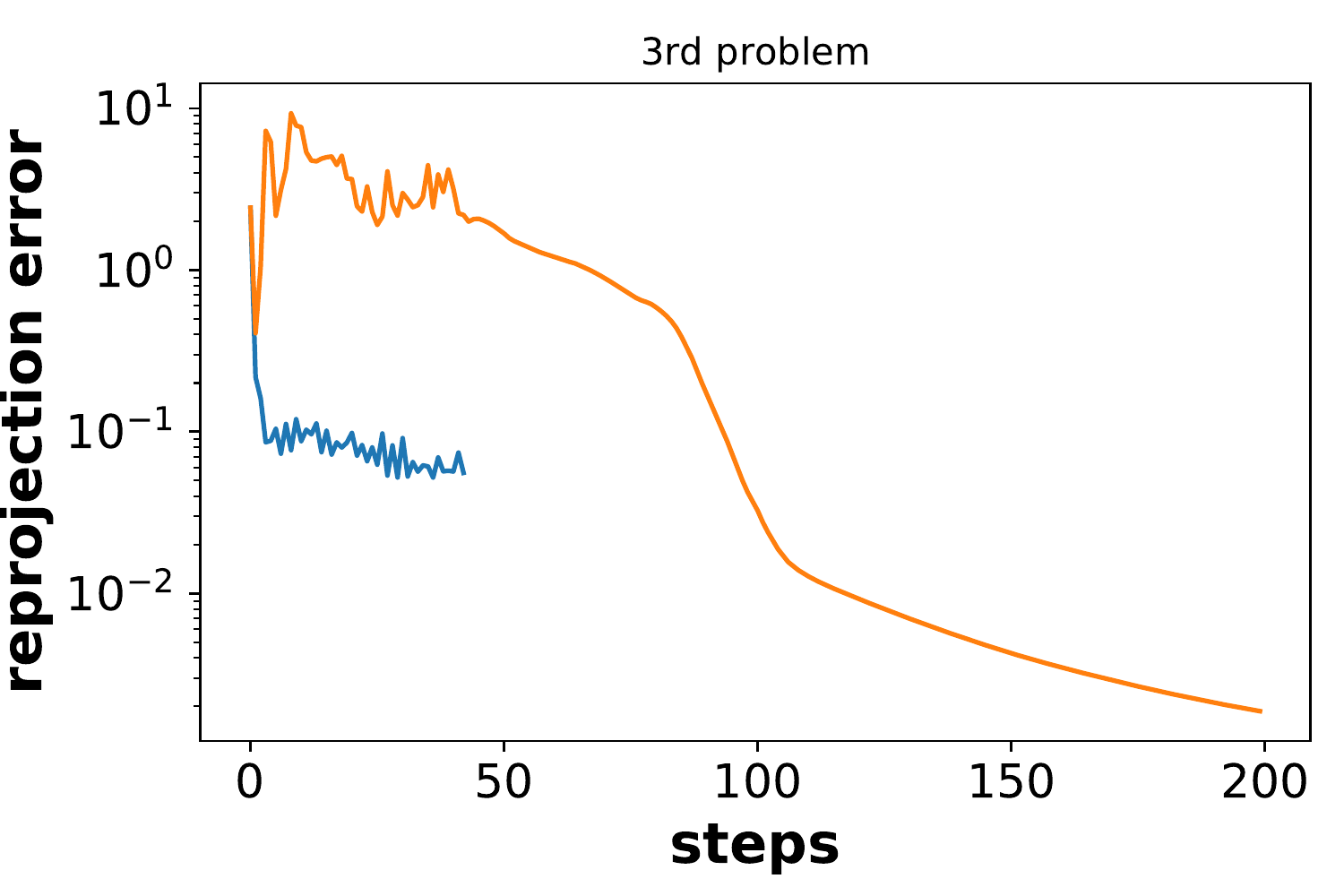}
\includegraphics[width = 0.45\textwidth]{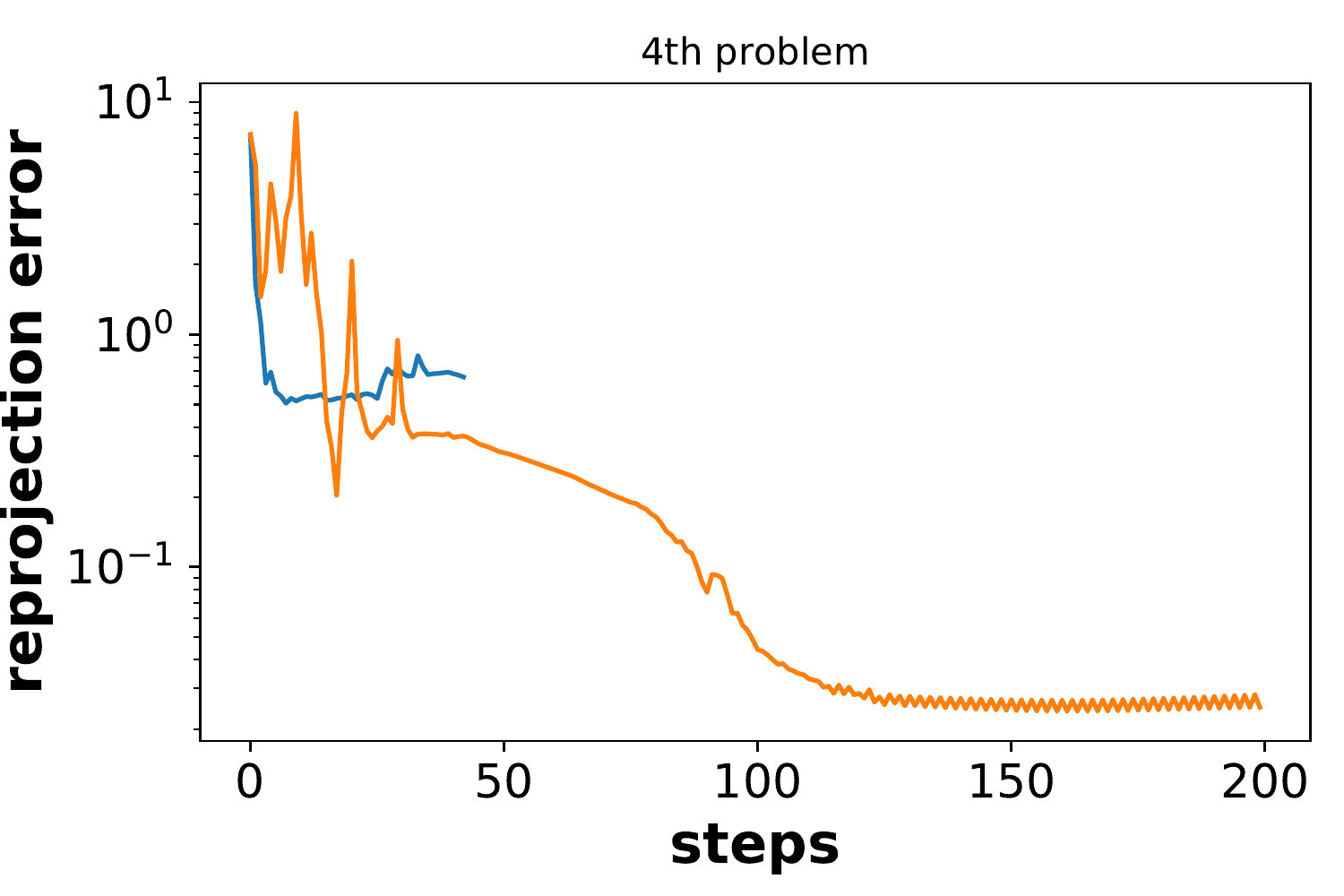}
\includegraphics[width = 0.45\textwidth]{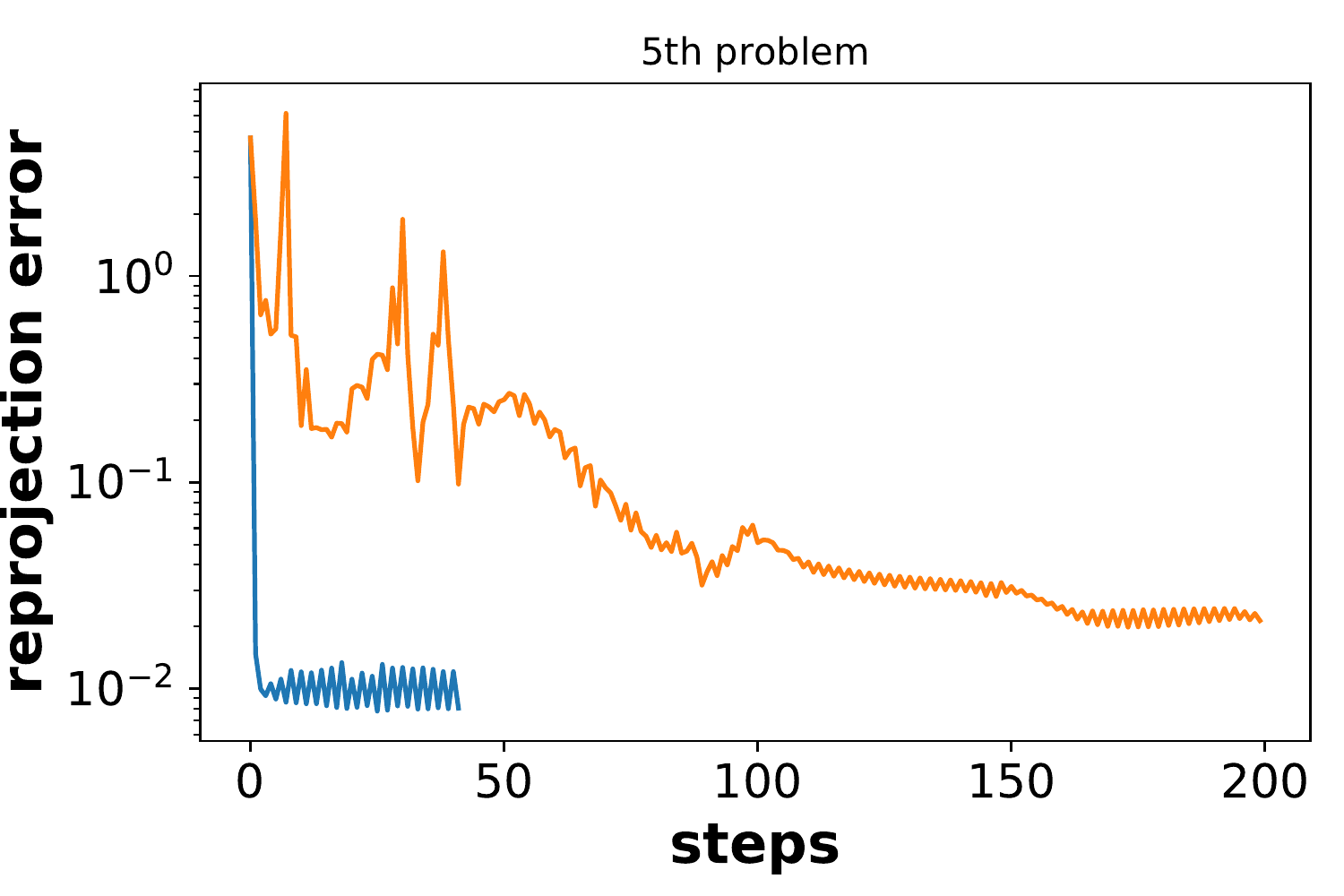}
\includegraphics[width = 0.45\textwidth]{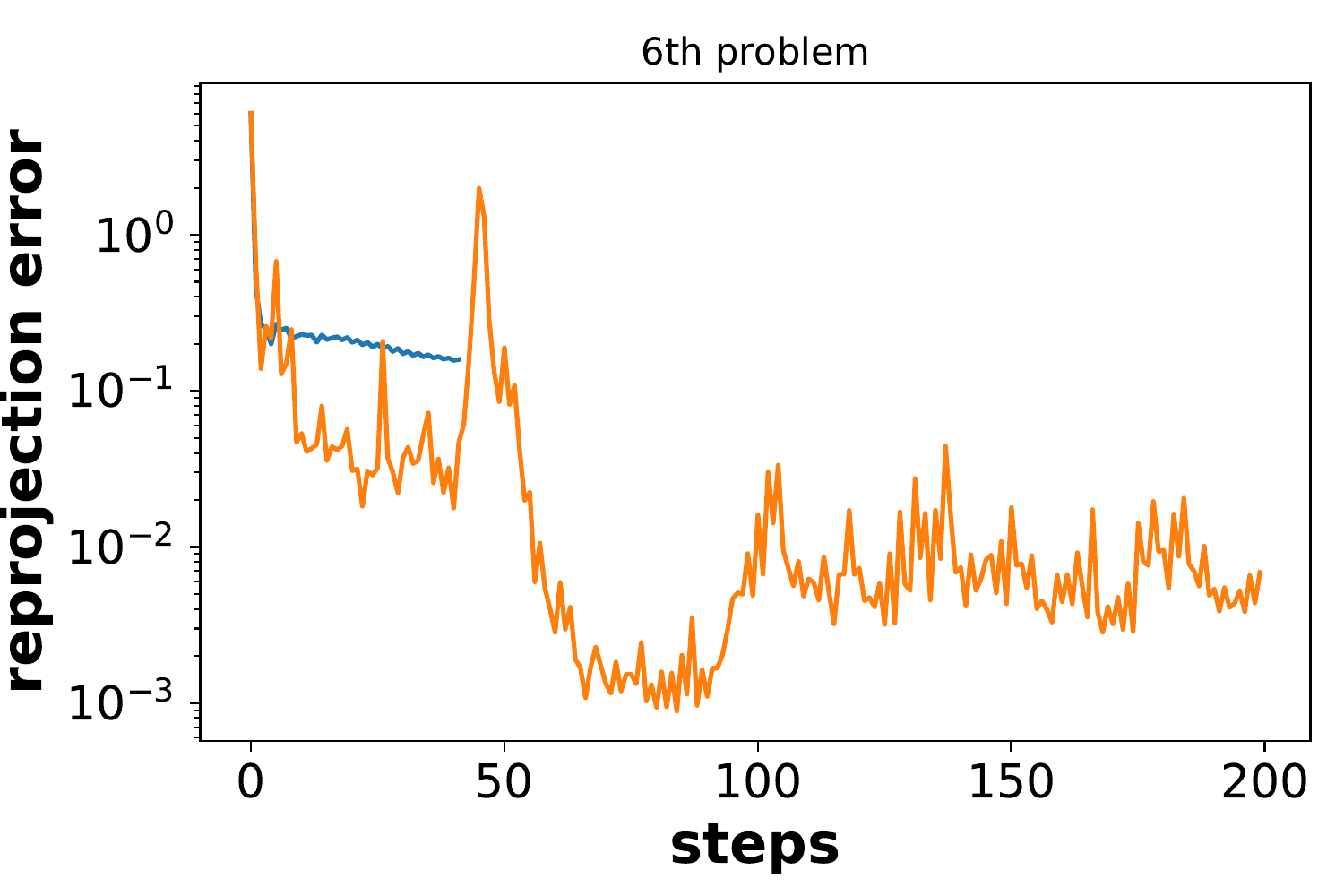}
\includegraphics[width = 0.45\textwidth]{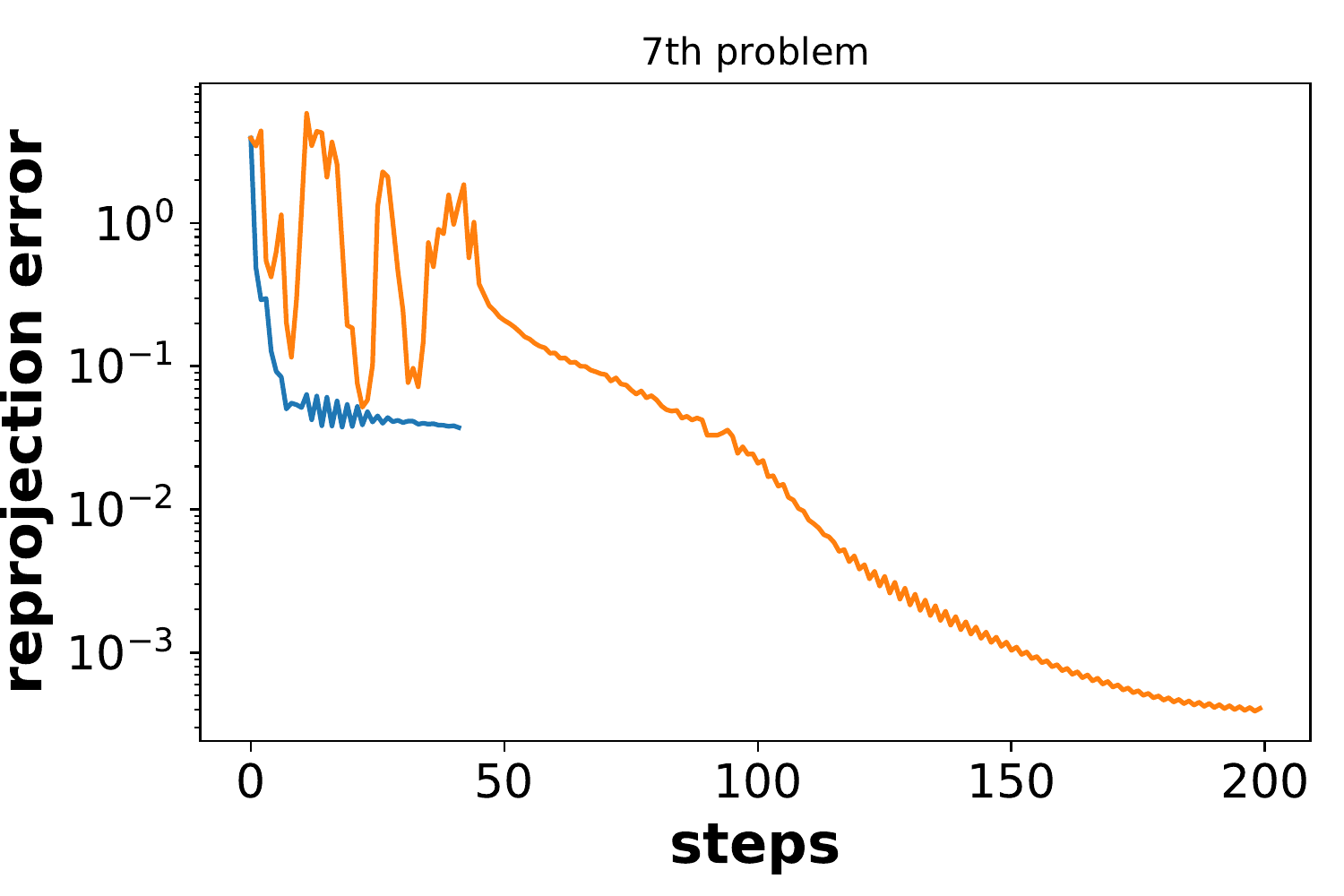}
\includegraphics[width = 0.45\textwidth]{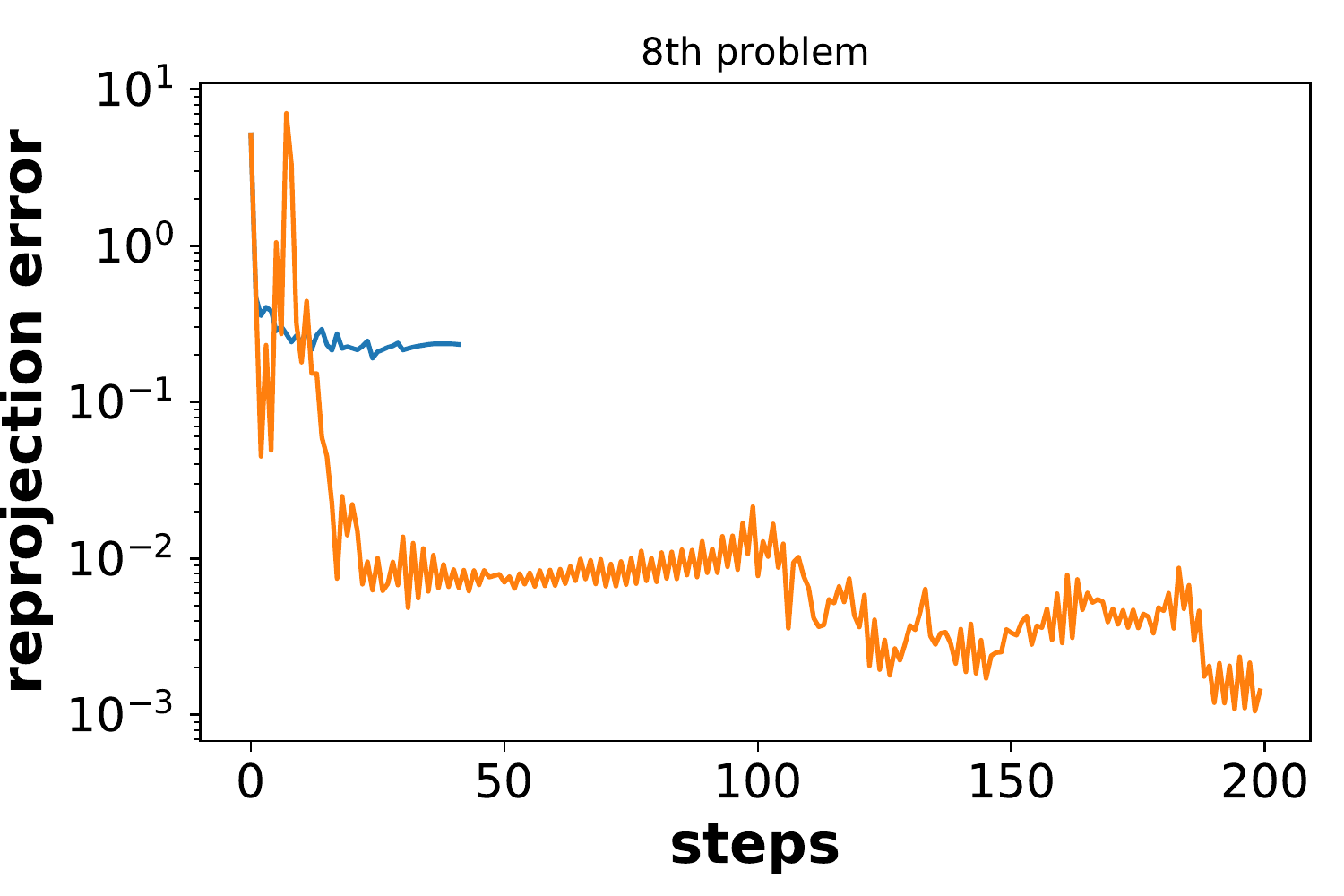}
\includegraphics[width = 0.45\textwidth]{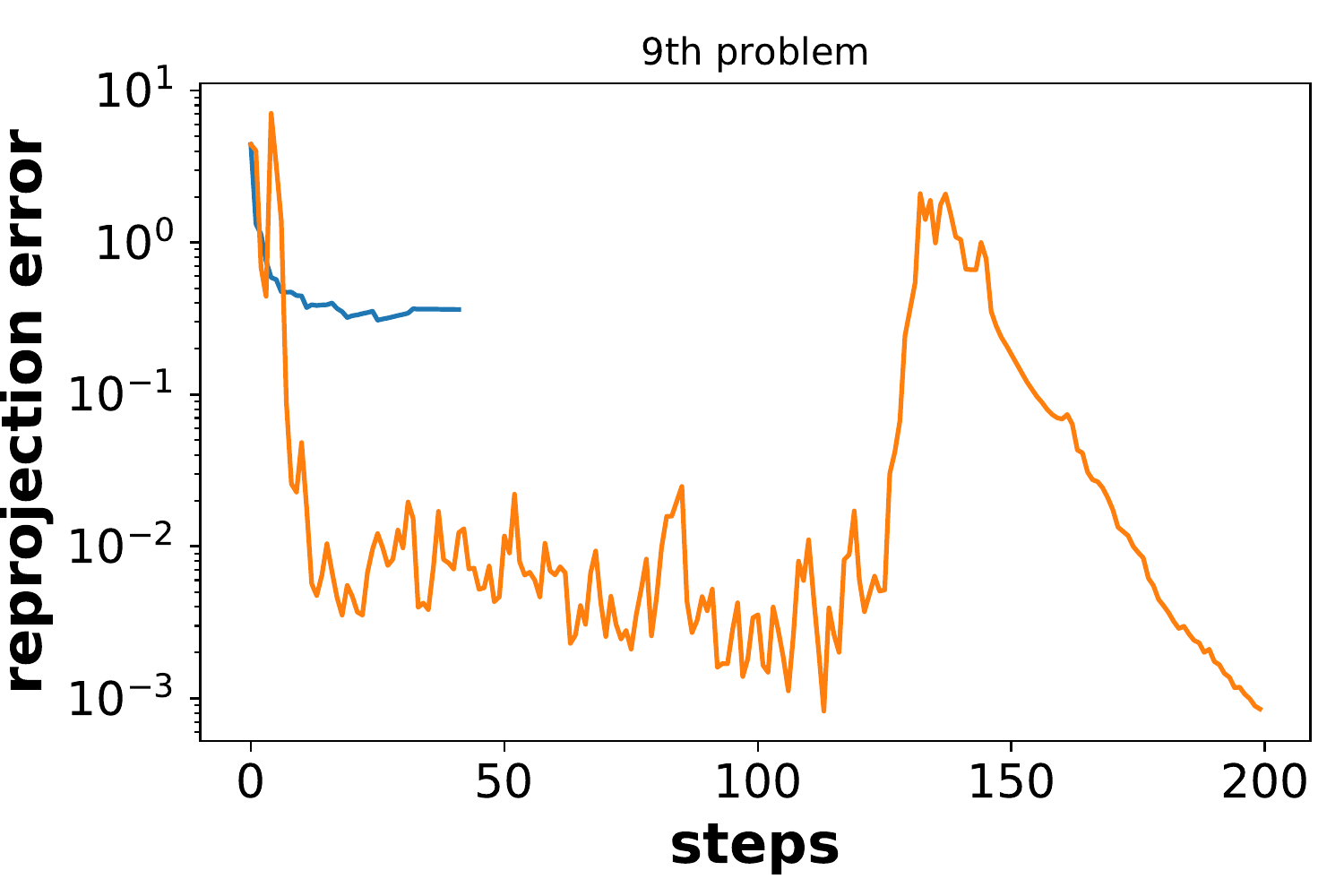}
\caption{Comparison of the sparse LMA to QLMA for the 9 random problems, where the QLMA has the parameters as in setup 1 and LMA as in setup 2 in Table 2 of the main text. The sparse LMA is denoted by the orange line and the QLMA by the blue line.}
\label{fig:6b}
\end{figure}

Here we show the cost convergence plots of all the problems under consideration and the results for the different setups. In the supplementary material is provided the optimization output for all problems. With a README.txt description on how to read the files. Each problem has been optimized with LMA with both setup 1 and setup 2 damping parameters, and with QLMA with only setup 1 damping parameters. One can see that for well constrained optimization landscapes like the 6th problem in Figs. \ref{fig:6a}, \ref{fig:6b} the QLMA offers no advantage to classical LMA, converging similarly at the beginning and being limited by the noise and approximations. On the other hand, challenging problems with difficult landscapes (like many of the problems under consideration in these examples) tend to take advantage of QLMA.



\section{HHL example}
\label{appendix:HHL}
Here we apply a simple analytical example of the HHL algorithm to explain each step of it. We additionally provide information regarding the relation with the HHL algorithm in general. Some steps become trivial due to the simplicity of the problem. For example, the decomposition of the unitary gates components for the QPE is not necessary. Therefore, no Trotter--Suzuki approximation is needed.
\newline
Let us assume we have the following problem,
\begin{align}
A x = b\Longleftrightarrow
\label{eq:linear}
    \begin{pmatrix}
    0&1\\
    1&0
    \end{pmatrix}
    \begin{pmatrix}
    y\\z
    \end{pmatrix}
    = \frac{1}{\sqrt{2}} \begin{pmatrix}
    1\\
    1
    \end{pmatrix}.
\end{align}
We could already make an educated guess that $x=y = 1/\sqrt{2}$ since $A^{-1}=A$. In general, one has to invert the matrix $A$ on the left, bring it onto the other side and solve it,
\begin{equation}
    \begin{pmatrix}
    y\\
    z
    \end{pmatrix}
    =    
    \begin{pmatrix}
    0&1\\
    1&0
    \end{pmatrix}^{-1}
    \frac{1}{\sqrt{2}} 
    \begin{pmatrix}
    1\\
    1
    \end{pmatrix},
\end{equation}
and classically the inversion of a general matrix is costly. How is this problem solved on a quantum computer using the HHL algorithm instead?

Quantum computers are built up by qubits, which are element of a two--dimensional Hilbert space that can be written in vector form as,
\begin{align}
\ket{0} = \begin{pmatrix}
1\\
0
\end{pmatrix},
\ket{1} = 
\begin{pmatrix}
0\\
1
\end{pmatrix}.
\end{align}
The quantum qubits are all connected by a tensor product.
\begin{figure}[h!]
\begin{quantikz}
\lstick{$\ket{0}$}&\qw\slice{1} &\qw\slice{2}&\qw\slice{3}&\qw\slice{4}&\gate{X}\gategroup[2,steps=1,style={dashed,
                   rounded corners, inner xsep=2pt},
                   background,label style={label position=above,anchor=
    north,yshift=0.5cm}]{{\sc INVERSION}}\slice{5}&\qw&\qw&\qw&\meter{}&\qw \\
\lstick{$\ket{0}$}&\qw &\gate{H}\gategroup[2,steps=3,style={dashed,
                   rounded corners, inner xsep=2pt},
                   background]{{\sc $QPE$}}&\ctrl{1}&\gate{H}&\octrl{-1}&\gate{H}\gategroup[2,steps=3,style={dashed,
                   rounded corners, inner xsep=2pt},
                   background]{{\sc $QPE^{\dagger}$}}&\ctrl{1}&\gate{H}&\meter{}&\qw\\
\lstick{$\ket{0}$}&\gate{H}&\qw&\gate{X}&\qw&\qw&\qw&\gate{X}&\qw&\qw&\qw
\end{quantikz}
\caption{Quantum computing operation can be represented as circuits, which can be seen here. The dashed lines just helps us as reference to explain each operational step. The explanation can be found in the text.}
\label{finalfigure}
\end{figure}
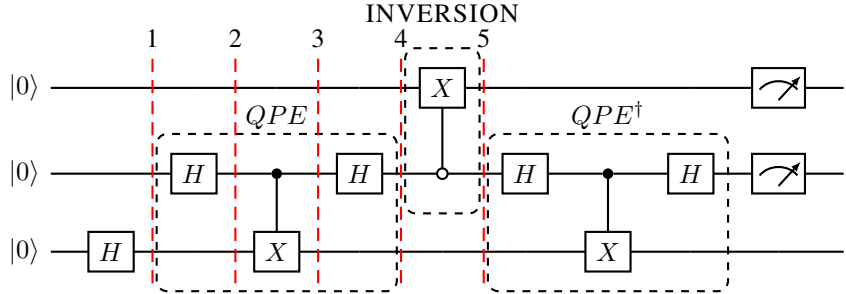

We set up a quantum circuit with three qubits, the first one is an ancilla register, which we will use later. Ancilla registers are additional qubits, needed to  implement algorithms. The second qubit is also an ancilla register which will be used for the QPE method and lastly the last qubit will encode the data.

We can apply operations onto such qubits (states) using unitary transformations called gates. For this example, we make mainly use of the quantum gate (operations) "Hadamard gate" and the flip gate X. 
Firstly, we are interested to encode our vector b in eq.(1) into the quantum computer. The Hadamard gate $H$ brings our state $\ket{0}$ to $1/\sqrt{2}(\ket{0}+\ket{1})$,
\begin{equation}
H = 
\frac{1}{\sqrt{2}}
    \begin{pmatrix}
    1&1\\
    1&-1
    \end{pmatrix}.
\end{equation}
We encode the data vector $\textbf{b}$ as $\ket{\psi} \equiv 1/\sqrt{2}(\ket{0}+\ket{1})$. Therefore, the total state of the computer becomes,
\begin{equation}
\label{colorequation}
 \begingroup\color{blue}\frac{1}{\sqrt{2}}\endgroup  (\ket{0}\ket{0}
\begingroup\color{blue}\ket{0}\endgroup
+\ket{0}\ket{0}\begingroup\color{blue}\ket{1}\endgroup) = \ket{0}\ket{0}\begingroup\color{blue}\ket{\psi}\endgroup,
\end{equation}
where the blue colored parts show the vector $\textbf{b}$, while the black $\ket{0}$s are unchanged ancilla qubits.
Then, we prepare our ancilla register of the QPE with a Hadamard gate. This results in the state that can be seen at step 2, 
\begin{equation}
    \frac{1}{\sqrt{2}}\ket{0}(\ket{0}+\ket{1})\ket{\psi}.
\end{equation}
Then we apply the control-X gate (CX).

The CX is a 2 qubit gate, and it is applied to 2 qubits simultaneously. The operation looks at the controlled qubit, which is seen in Fig. \ref{finalfigure} by a black dot. If it is $\ket{1}$ it applies an operation X onto the target register and performs a flip, which can be seen by a box with an X inside. In other words, X gate switches $\ket{1}$ to $\ket{0}$ and vice versa and the CX is given as,
\begin{equation}
\textrm{CX} = \begin{pmatrix}
1&0&0&0\\
0&1&0&0\\
0&0&0&1\\
0&0&1&0
\end{pmatrix}.
\end{equation}
Our CX gate is in the general a time evolution of our matrix and is a composition of arbitrary unitary operations, which can be carried on the quantum computer, as described in Eq.(4) in the main text. To apply the QPE we evaluate the time evolution in eq.(4) in the main text. $e^{-iAt}$, where A is Hermitian. $e^{-iAt}$ is unitary and can be expressed in gates. In our case the matrix A corresponds to the X gate and therefore our $e^{-iAt}$ is also implemented by an X gate. In general, the matrix $A$ and it's evolution $e^{-iAt}$ rarely have the same matrix representation and for this reason are decomposed and approximated using Trotter--Suzuki. The CX gate implements the Hamiltonian evolution $e^{iAt}$ if our control qubit is 1. A general evolution results in a the state at step 3,
\begin{equation}
    \frac{1}{\sqrt{2}}\ket{0}(\ket{0}+e^{2i\pi 0.\phi_1}\ket{1})\ket{\psi}.
\end{equation}
Here it can be seen that $\phi_1$ has to be 0 for the CX gate to obtain the correct results. This is because our solution is exact. Imagine we would have a non exact solution and the true value would lay between 0 and 1, then our $\phi_1$ would share a probability of being either in 0 and 1. With more ancilla qubits reserved for subsequent QPE measures it would be possible to get closer to an exact solution.

Notice that this control-gate can be a lot more complex, it can be that it is a composition of non commuting terms. In this case it would be necessary to apply the Trotter-Suzuki method, which is described in the main part of the paper.
Then after step 3 we apply another Hadamard gate, which acts as the inverse quantum Fourier transform, which makes it possible to write the exponent into a state. Additionally, $\ket{\psi}$ is a composition of the eigenvectors $\sum_j\ket{u_j}$ so we may straight write it as $\sum_j\ket{u_j}$.
\begin{equation}
   \sum_j \frac{1}{2}((1+e^{2\pi i 0.\phi_1})\ket{0}+(1-e^{2\pi i 0.\phi_1})\ket{1})\ket{0}\ket{u_j}.
\end{equation}
It can be noticed, as previously, that if $\phi_1=0$ then this results in,
\begin{equation}
     \frac{1}{2}((1+1)\ket{0}+(1-1)\ket{1})\ket{0}\ket{\psi} =\sum_j \lambda_j\ket{0}\ket{0}\ket{u_j},
\end{equation}
where $\lambda$ are the actual eigenvalues for the matrix for eigenvectors, which are both 1. 

We can then take a further look into the final crucial stage of the HHL, which is the matrix inversion. The first qubit, which has been untouched so far, will be manipulated. In this problem this is again simplified because the inverse of $A$ is itself and due to $\lambda_j=1$, we have $1/\lambda_j$=$\lambda_j$. In general, we wish to move our first qubit to obtain,
\begin{equation}
    \sum_j\sqrt{1-\frac{C^2}{\lambda_j^2}}\ket{0}+\frac{C}{\lambda_j}\ket{1},
\end{equation}
where C is just a scalar between 0 and 1 obtained in the inversion, where in our case it is 1. $\lambda_j$ are generally the different eigenvalues, which in our case are both 1. Then we are interested in measuring $\ket{1}$, which holds the information of the inverse eigenvalues. Therefore the first qubit is being prepared with an Hadamard gate and then the matrix inversion is applied so that we result in the equation above.

In our case we can simply say if our second qubit is in the state $\ket{0}$, this is indicated by the circle between step 4 and 5 on the second qubit, which is the state with the information of our eigenvalues, then we rotate $\ket{0}$ to $\ket{1}$. 
At step 5, we therefore have.
\begin{equation}
    \sum_j \frac{1}{\lambda_j}\ket{0}\ket{1}\ket{u_j}.
\end{equation}
Then we are supposed to apply the QPE backwards to get rid of the QPE state $\ket{0}$ and the coefficient before, which in our case is only 1, this state then can be measured which concludes the circuit in Fig. \ref{finalfigure},
\begin{equation}
\label{final}
   \sum_j \lambda_j^{-1}\ket{u_j} = \frac{1}{\sqrt{2}}\left(\ket{0}+\ket{1}\right).
\end{equation}
And we can write $\ket{x} = \sum_j\lambda^{-1}_j\ket{u_j}$, which is exactly Eq.(\ref{final}) and solution to Eq. (\ref{eq:linear}). It may be noted again that $b$ in Eq.(\ref{eq:linear}) is decomposed into $\ket{u_j}$ on the quantum computer. Our final state encodes the solution desired.
Now the algorithm can be run an arbitrary number of times and the last qubit is measured this would give us a probability of measuring the probability 1/2 for each $\ket{0}$ and $\ket{1}$. The more often we run it the higher statistical probability we get to 1/2. Combining the results of the measurements gives us the full solution,
\begin{equation}
x = \begin{pmatrix}
y\\
z
\end{pmatrix}
=
\frac{1}{\sqrt{2}}\begin{pmatrix}
1\\
1
\end{pmatrix}.
\end{equation}

\bibliography{references.bib}
\bibliographystyle{unsrt}
\end{document}